\documentclass[10pt,twocolumn,letterpaper]{article}

\usepackage[pagenumbers]{cvpr} 

\definecolor{cvprblue}{rgb}{0.21,0.49,0.74}
\usepackage[pagebackref,breaklinks,colorlinks,allcolors=cvprblue]{hyperref}

\usepackage{booktabs}   
\usepackage{multirow}
\usepackage{makecell}
\usepackage{amsmath}
\usepackage[table]{xcolor} 
\usepackage[T1]{fontenc}

\usepackage[skins,breakable]{tcolorbox}
\tcbuselibrary{listings}
\usepackage{tcolorbox}
\usepackage{mdframed}
\usepackage{marvosym}
\usepackage{colortbl}
\usepackage{enumitem}
\usepackage{fancyvrb}
\usepackage{algorithm}
\usepackage{algorithmic}

\def\paperID{*****} 
\def\confName{CVPR}
\def\confYear{2026}

\title{CORE: \underline{Co}de-based Inve\underline{r}se S\underline{e}lf-Training Framework \\ with Graph Expansion for Virtual Agents}

\author{
Keyu Wang\textsuperscript{1}\thanks{Equal Contribution.}~~
Bingchen Miao\textsuperscript{1}\footnotemark[1]~~
Wendong Bu\textsuperscript{1}~~
Yu Wu\textsuperscript{2}~~
Juncheng Li\textsuperscript{1}\thanks{Corresponding Author.}\\
Shengyu Zhang\textsuperscript{1}~~
Wenqiao Zhang\textsuperscript{1}~~
Siliang Tang\textsuperscript{1}~~
Jun Xiao\textsuperscript{1}~~
Yueting Zhuang\textsuperscript{1}\\
{\small \textsuperscript{1}Zhejiang University~~
\textsuperscript{2}Wuhan University}\\
{\tt\small \{keyuwang, miaobingchen23, wendongbu, junchengli, sy\_zhang,}\\
{\tt\small wenqiaozhang, siliang, junx, yzhuang\}@zju.edu.cn, wuyucs@whu.edu.cn}
}

\begin{document}
\maketitle
\begin{abstract}
The development of Multimodal Virtual Agents has made significant progress through the integration of Multimodal Large Language Models. However, mainstream training paradigms face key challenges: Behavior Cloning is simple and effective through imitation but suffers from low behavioral diversity, while Reinforcement Learning is capable of discovering novel strategies through exploration but heavily relies on manually designed reward functions. To address the conflict between these two methods, we present \textbf{CORE}, a \underline{\textbf{Co}}de-based Inve\underline{\textbf{r}}se S\underline{\textbf{e}}lf-Training Framework with Graph Expansion that bridges imitation and exploration, offering a novel training framework that promotes behavioral diversity while eliminating the reliance on manually reward design. Specifically, we introduce \textbf{Semantic Code Abstraction} to automatically infers reward functions from expert demonstrations without manual design. The inferred reward function, referred to as the Label Function, is executable code that verifies one key step within a task. Building on this, we propose \textbf{Strategy Graph Expansion} to enhance in-domain behavioral diversity, which constructs a multi-path graph called Strategy Graph that captures diverse valid solutions beyond expert demonstrations. Furthermore, we introduce \textbf{Trajectory-Guided Extrapolation}, which enriches out-of-domain behavioral diversity by utilizing both successful and failed trajectories to expand the task space. Experiments on Web and Android platforms demonstrate that CORE significantly improves both overall performance and generalization, highlighting its potential as a robust and generalizable training paradigm for building powerful virtual agents.
\end{abstract}
\vspace{-10pt}    
\section{Introduction}
\label{sec:intro}

\begin{figure}[t]
\centering
\includegraphics[width=0.45\textwidth]{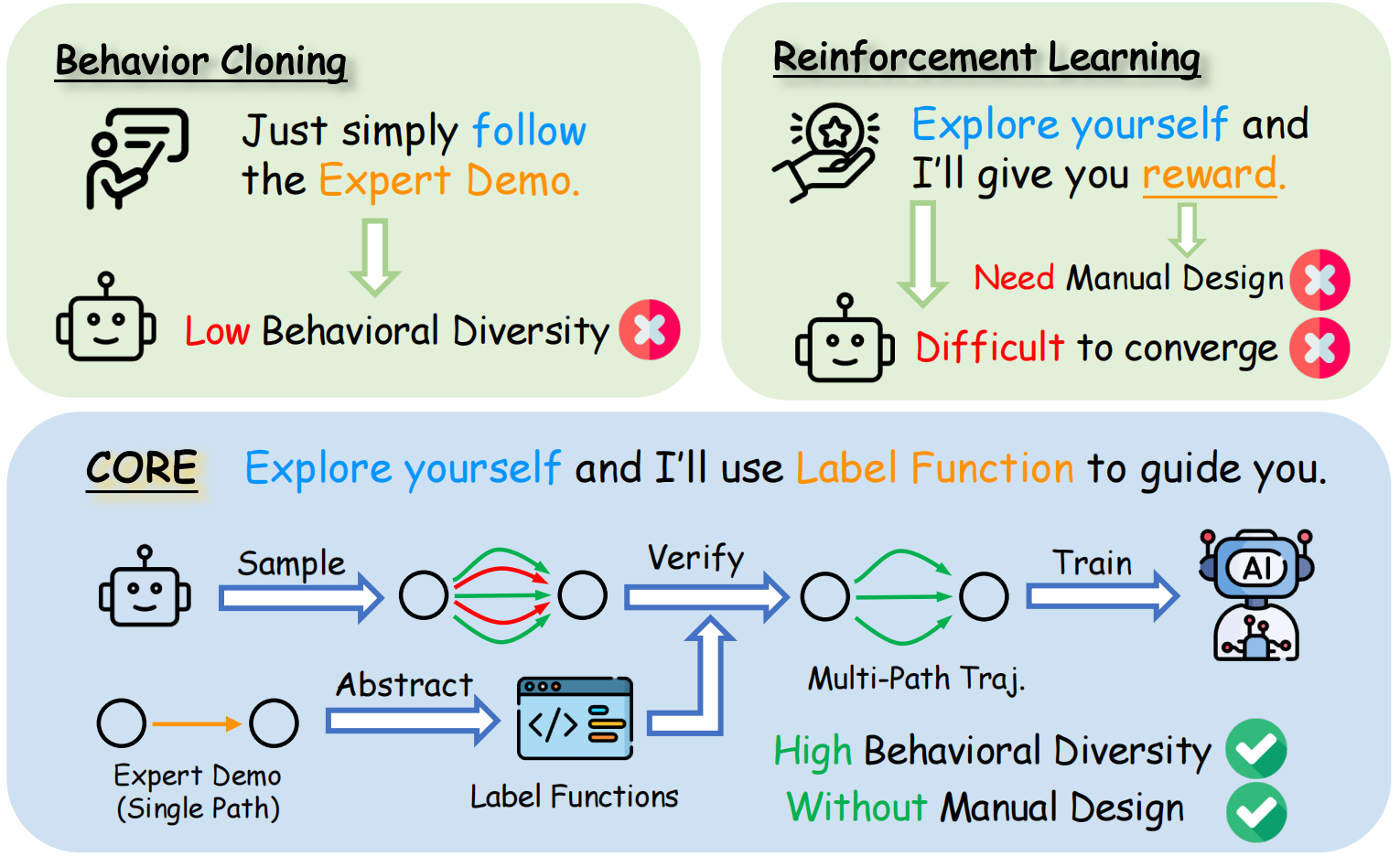} 
\caption{Top: Limitations of Behavior Cloning and Reinforcement Learning. Bottom: Illustration of CORE, our self-training framework that bridges imitation and exploration. }
\label{fig1}
\vspace{-15pt}
\end{figure}

Multimodal Virtual Agents~\cite{lin2025showui, xu2024aguvis}, empowered by the advanced capabilities of Multimodal Large Language Models ~\cite{zhang2024mm, yin2024survey, li2023fine}, are capable of taking multimodal signals such as GUI screenshots and natural language instructions as inputs, enabling them to autonomously plan and execute actions like controlling mobile apps~\cite{wang2025mobile}, clicking interface elements~\cite{gou2024navigating}, and browsing webpages~\cite{shen2024scribeagent} within complex digital environments~\cite{pan2024webcanvas, xie2024osworld, bu2025limits}. Currently, the mainstream approaches for training agents can be broadly categorized into two types: \textbf{Behavior Cloning}~\cite{guan2024explainable}, which learns policies by imitating expert demonstrations, and \textbf{Reinforcement Learning}~\cite{ibrahim2024comprehensive}, which optimizes policies through interaction with the environment based on reward signals.

However, these two types of training methodologies each come with their own critical limitations: \textbf{1) Low Behavioral Diversity in Behavior Cloning:} Behavior Cloning leverages expert demonstrations to learn action policies in a supervised manner, making it a simple, effective and stable approach. However, this imitation-based approach suffers from low behavioral diversity. Achieving a goal often involves multiple valid paths, yet Behavior Cloning tends to overfit to the specific sequences provided by experts, leading to limited generalization in unseen scenarios~\cite{wan2024quality, ankile2024imitation, li2025structure}. Moreover, collecting high-quality expert demonstrations is costly and difficult to scale up, ultimately restricting the approach's applicability~\cite{liang2024constrained}. \textbf{2) Reliance on Manual Reward Design in Reinforcement Learning:} Reinforcement Learning optimizes agent behavior by maximizing cumulative reward through environment interaction, enabling the discovery of novel strategies beyond human.  This exploration-based paradigm, despite its potential, heavily relies on manually crafted reward functions, which can be time-consuming and may lead to suboptimal behaviors if poorly designed~\cite{wang2025m3hf}. Furthermore, it's challenging to craft effective reward functions, especially process-based ones, leading to sparse rewards that make it difficult for the model to converge~\cite{ibrahim2024comprehensive}. \textbf{In summary}, the trade-off between stability through imitation and generalization through exploration underscore the need for a more robust and generalizable training framework for virtual agents.

To address these challenges, we propose \textbf{CORE}, a Code-based Inverse Self-Training Framework with Graph Expansion for Virtual Agents, offering a novel training framework that bridges imitation and exploration, achieving high diversity while eliminating the need for hand-crafted reward functions, as illustrated in Figure~\ref{fig1}. Specifically, \textbf{1)} we introduce \textbf{Semantic Code Abstraction}, a method that inversely infers key process reward functions from expert demonstrations without manual design. This reward function, termed the Label Function, is executable code that verifies whether the agent completes one specific key step within a task. To generate Label Functions, we transform expert demonstrations into semantic descriptions, identify action steps relevant to the task goal as key steps, and synthesize Label Functions based on these key steps. Building on this, \textbf{2)} we propose \textbf{Strategy Graph Expansion}, a method designed to enhance in-domain behavioral diversity by iteratively exploring novel strategies beyond expert data. For each task, Label Functions derived from expert demonstrations act as reward signals to classify trajectories sampled by the agent itself into three types: trajectories that fully pass, partially pass, or fail all Label Functions. As partially passed trajectories may reveal valid solutions distinct from the expert sequence, we infer new Label Functions from them and integrate these into the existing sequence, thereby constructing and expanding a multi-path graph referred to as the Strategy Graph. Once constructed, the graph serves as a verifier to evaluate newly sampled trajectories, accepting diverse trajectories aligned with different valid solutions and thereby enriching behavioral diversity iteratively. On this basis, \textbf{CORE} combines the stability of Behavior Cloning with the generalization of Reinforcement Learning, avoiding low behavioral diversity while eliminating the need for hand-crafted reward functions.

Furthermore, we introduce \textbf{Trajectory-Guided Extrapolation}, a mechanism designed to enrich out-of-domain behavioral diversity by leveraging both successful and failed trajectories. \textbf{1)} Leveraging successful trajectories, the agent autonomously discovers new solvable tasks absent in expert demonstrations but achievable under its current policy. \textbf{2)} Simultaneously, by leveraging failed trajectories, we treat them as successful under better-aligned``task intents" (the specific goals or objectives of the task) by inferring plausible novel intents. Overall, this method enables the agent to utilize all trajectories effectively, thus expanding task space beyond expert data. Together with CORE, this approach further enhances overall diversity, resulting in a unified and diverse self-training framework.


Extensive experiments demonstrate that \textbf{CORE} significantly boosts the overall performance and generalization of virtual agents. We evaluate our framework across interactive environments on both Web and Android platforms. Experiment results show that it consistently outperforms Behavior Cloning, Reinforcement Learning and prior self-training methods, achieves steady improvements over multiple training iterations, and exhibits strong generalization to unseen tasks. These results highlight the effectiveness and robustness of our framework.


Our contributions can be summarized as follows:

\begin{itemize}
    \item We introduce \textbf{CORE}, a Code-based Inverse Self-Training Framework that proposes the \textbf{Semantic Code Abstraction} and the \textbf{Strategy Graph} to capture diverse solution paths. This rule-based framework overcomes the core conflict between imitation and exploration.
    \item We introduce \textbf{Trajectory-Guided Extrapolation}, which further enriches behavioral diversity by expanding the task space, complementing \textbf{CORE} as a unified self-training framework to enhance generalization.
    \item Experiments demonstrate that our framework is a robust and generalizable training paradigm for building powerful virtual agents.
\end{itemize}
\begin{figure*}[t]
\vspace{-10pt}
\centering
\includegraphics[width=1\textwidth]{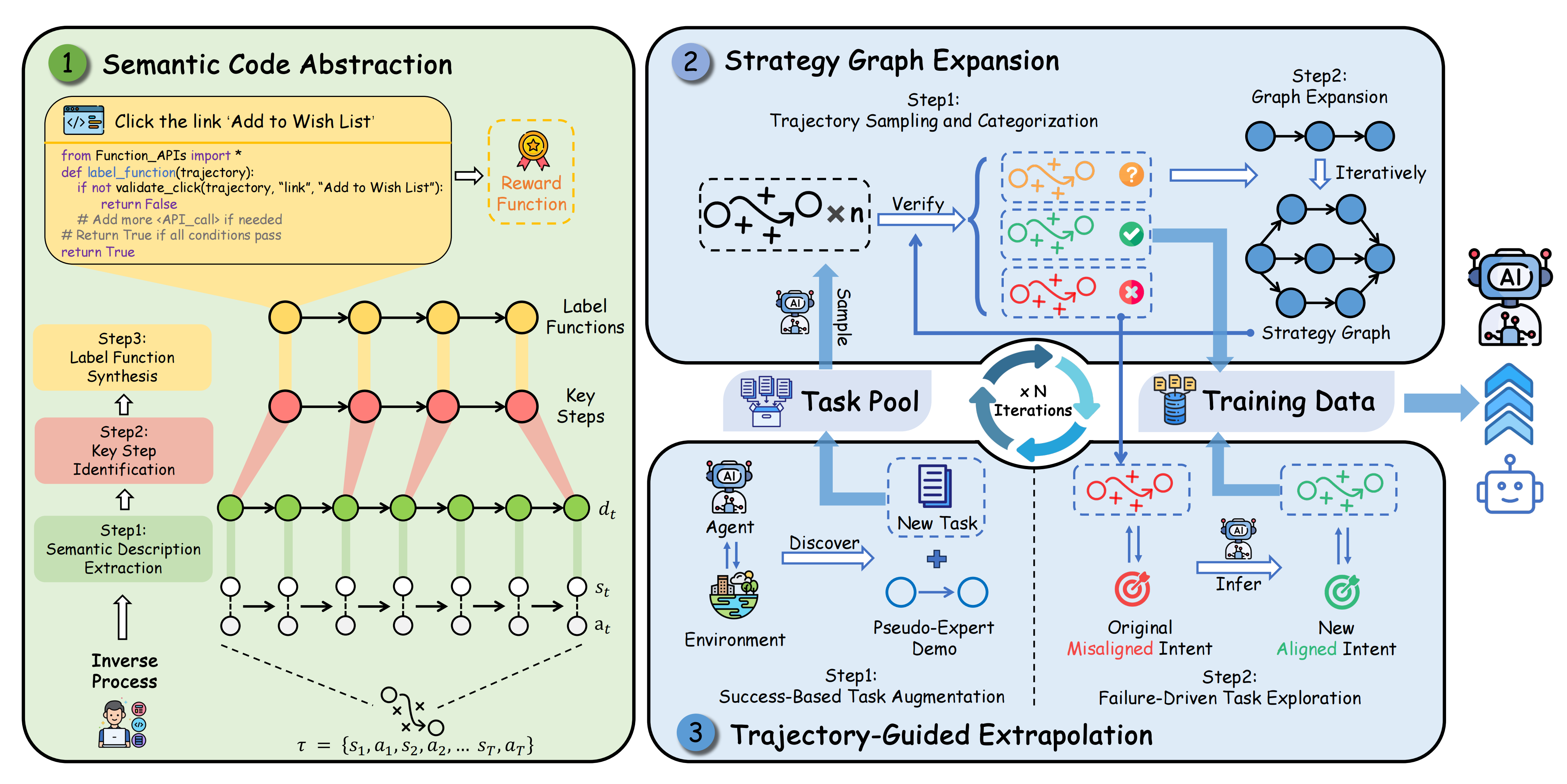}
\caption{\textbf{Overview of CORE}: A Code-based Inverse Self-Training Framework that derives executable reward functions from expert demonstrations via Semantic Code Abstraction, enhances in-domain diversity through multi-path Strategy Graph Expansion, and enriches out-of-domain diversity by recycling trajectories with Trajectory-Guided Extrapolation, forming a iterative self-improving pipeline for training robust and generalizable Multimodal Virtual Agents.}
\label{fig:framework_structure}
\vspace{-10pt}
\end{figure*}

\section{Related Work}
\label{sec:related_work}

\subsection{Behavior Cloning \& Reinforcement Learning}

Behavior Cloning (BC)~\cite{torabi2018behavioralcloningobservation} leverages expert demonstrations to learn action policies via supervised learning, achieving high initial task success in agent domains like web navigation~\cite{zhou2023webarena} and robotics~\cite{Bousmalis2023RoboCat}. However, Behavior Cloning's imitation paradigm inherently limits behavioral diversity and generalization~\cite{sodhi2024stepstackedllmpolicies, srinivasan2025webnavintelligentagentvoicecontrolled, li2023variational}. Reinforcement Learning (RL)~\cite{RL2018} addresses diversity via reward-driven exploration but depends heavily on manual reward design, which can lead to sparse rewards and unintended behaviors~\cite{le2022coderl}. For instance, DigiRL~\cite{bai2024digirl}, while advancing agent capabilities in realistic digital environments, still requires careful engineering of reward functions to guide exploration effectively. Our work bridges this gap by automating reward synthesis and integrating diverse exploration, eliminating manual design while enhancing behavioral diversity.

\subsection{Self-Training for Multimodal Virtual Agents}

Multimodal Virtual Agents process inputs like GUI screenshots and language instructions, enabling interaction with complex environments~\cite{gao2024generalistvirtualagentssurvey, miao2025boostingvirtualagentlearning}. Self-Training~\cite{huang-etal-2023-large, yuan2023scaling, dong2024self, song2024trial} has emerged as a promising method for data exploration, filtering, and refinement with limited supervision, creating an iterative exploration-feedback-optimization cycle. Specifically, some approaches focus on data curation and augmentation. For example, Self-Improve~\cite{patel2024large} filters in-domain trajectories and synthesizes diverse out-of-domain data to enhance the agent's performance. Similarly, BAGEL~\cite{murty2024bagel} improves agents via iterative exploration and labeling. However, existing self-training approaches for virtual agents are either limited to a single platform or fail to iteratively refine agent behavior. Our framework addresses this by introducing a robust and generalizable training paradigm for iteratively refining generalist virtual agents.
\section{Method}
\label{sec:method}

In this section, we introduce the detailed methodology of \textbf{CORE}, a Code-based Inverse Self-Training Framework with Graph Expansion for Virtual Agents. As shown in Figure~\ref{fig:framework_structure}, our framework leverages \textbf{Semantic Code Abstraction} (Section~\ref{3.1}) to automatically derive reward functions from expert demonstrations, eliminating the need for manual reward design. These functions enable fine-grained evaluation of trajectories, guiding \textbf{Strategy Graph Expansion} (Section~\ref{3.2}), which enriches in-domain diversity through multi-path expansion in the strategy graph, and \textbf{Trajectory-Guided Extrapolation} (Section~\ref{3.3}), which extends task space by recycling both successful and failed trajectories. Together, these components form a self-improving pipeline that boosts performance for Multimodal Virtual Agents. See Appendix~\ref{Pseudocode} for detailed pseudocode of each component and the overall pipeline.

\begin{figure*}[t]
\centering
\includegraphics[width=0.9\textwidth]{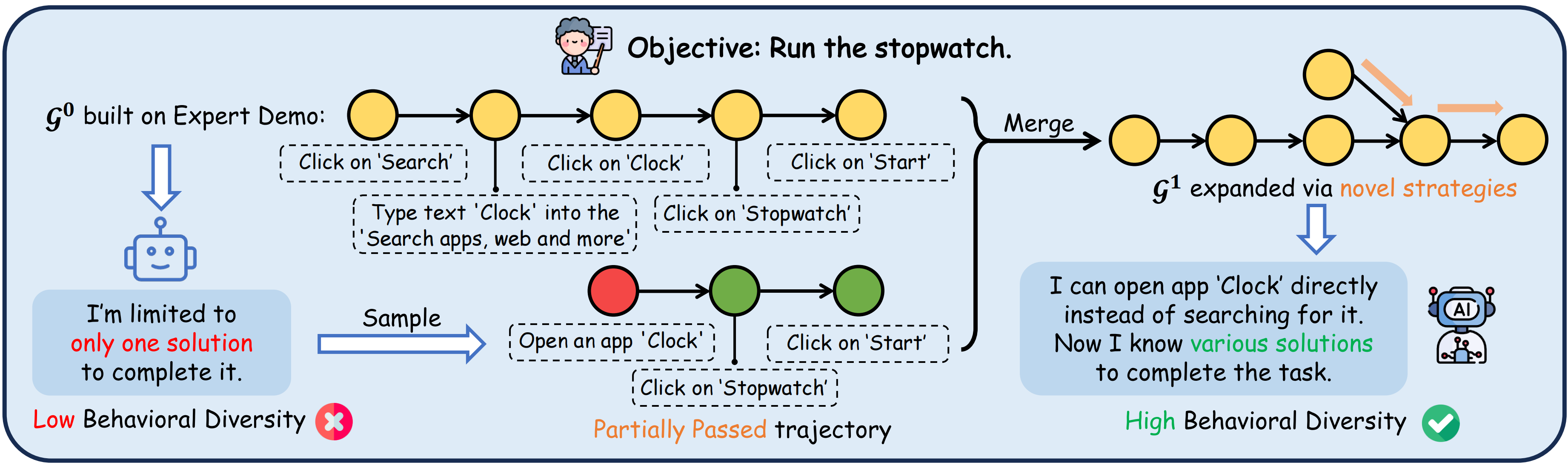}
\caption{An example demonstrating how Strategy Graph Expansion grows the graph and enriches behavioral diversity.}
\label{fig:method2_example}
\end{figure*}

\subsection{Semantic Code Abstraction}
\label{3.1}

As Reinforcement Learning heavily relies on manually crafted reward functions, Semantic Code Abstraction aims to inversely infer executable reward functions, termed Label Functions, from expert demonstrations without manual design. To generate them, we first translate expert demonstrations into semantic descriptions, then identify key steps, and synthesize Label Functions based on these key steps.

\subsubsection{Semantic Description Extraction.}
\label{3.1.1}

To begin, we transform expert demonstrations into semantic descriptions that capture the context of each action. Given a trajectory 
\begin{equation}
    \tau = \{(s_1, a_1), (s_2, a_2), \dots, (s_T, a_T)\},
\end{equation}
where \( s_t \) represents the environment state (e.g., GUI screenshots or Accessibility Tree) and \( a_t \) denotes the action at time \( t \), we define a semantic abstraction function:
\begin{equation}
    \mathcal{D}: (s_t, a_t) \mapsto d_t,
\end{equation}
that maps each pair to a semantic description \( d_t \), such that the high-level abstraction of the trajectory is:
\begin{equation}
    \mathcal{D}(\tau) = \{d_1, d_2, \dots, d_T\}.
\end{equation}

The function $\mathcal{D}$ extracts the referent object of $a_t$ from $s_t$, and formats it into natural language using a predefined mapping template since the action space in each environment is fixed. For example, from $a_t = \text{Click [42]}$ and $s_t$ containing ``[42] [A] [Add to Wish List]'', we obtain \( d_t = \) ``Click the link `Add to Wish List'''. This enables our model to understand the trajectory at a semantic level by utilizing its powerful language understanding ability~\cite{wei2022emergent}.

\subsubsection{Key Step Identification.}
\label{3.1.2}

In this step, we aim to identify the \textbf{key steps} in a trajectory, defined as actions directly relevant to the task goal and advancing its progress.

To formally capture this process, we define a mapping:
\begin{equation}
    \mathcal{K}: \left( \mathcal{D}(\tau), g \right) \mapsto \mathcal{D}_{\text{key}} \subseteq \mathcal{D}(\tau),
\end{equation}
where $\mathcal{D}(\tau)$ is the full sequence of semantic descriptions in trajectory $\tau$ as defined in Semantic Description Extraction (Section~\ref{3.1.1}), $g$ is the specified task goal, and $\mathcal{D}_{\text{key}}$ represents the key steps.

To identify $\mathcal{D}_{\text{key}}$, our model applies the mapping $\mathcal{K}$ by using a prompt (details in Appendix~\ref{Prompt for Key Step Identification}) that includes $\mathcal{D}(\tau)$ and goal $g$. Conditioned on this prompt, the model assesses the semantic relevance of each action and selects only those steps that substantially contribute to the task goal. 

\subsubsection{Label Function Synthesis.}
\label{3.1.3}

To evaluate whether an agent has successfully completed key steps in a task, we propose \textbf{Label Functions} which are executable Python code snippets that act as key process reward functions. Each Label Function takes a trajectory $\tau$ as input and outputs a binary value indicating whether a particular key step has been successfully executed. Formally, each Label Function is defined as a binary classifier:
\begin{equation}
    L_k: \tau \mapsto \{0, 1\},
\end{equation}
where $L_k$ corresponds to the $k$-th key step $d_k \in \mathcal{D}_{\text{key}}$. The complete set of Label Functions for a task is:
\begin{equation}
    \mathcal{L} = \{L_1, L_2, \dots, L_K\}.
\end{equation}

To generate these Label Functions, we define a synthesis function mapping semantic descriptions to executable code:
\begin{equation}
    \mathcal{S}: (d_k, \mathcal{A}) \mapsto L_k,
\end{equation}
where $d_k \in \mathcal{D}_{\text{key}}$ is the semantic description of the $k$-th key step, $\mathcal{A}$ is a predefined API set, and $L_k$ is the generated Label Function. The synthesis process $\mathcal{S}$ leverages a code generation model to translate $d_k$ into executable code by selecting and composing relevant APIs from $\mathcal{A}$ based on the action and context described in $d_k$. See the Appendix~\ref{Prompt for Label Function Synthesis} for detailed prompts. This process ensures that each task, paired with an expert demonstration trajectory $\tau$, can be decomposed into a sequence of Label Functions $\mathcal{L}$, enabling automatic high-level evaluation of agent behavior.

\subsection{Strategy Graph Expansion}
\label{3.2}
As Behavior Cloning suffer from low behavioral diversity, Strategy Graph Expansion aims to address this issue by constructing a multi-path graph, termed the Strategy Graph. We iteratively sample trajectories, categorize them, and expand Strategy Graph to refine the policy and enhance in-domain behavioral diversity over multiple rounds. An example of this approach is illustrated in Figure~\ref{fig:method2_example}.

The Strategy Graph $\mathcal{G} = (\mathcal{V}, \mathcal{E})$ is a directed acyclic graph ~\cite{xu2024crab} where vertices $\mathcal{V}$ represent Label Functions, and edges $\mathcal{E}$ capture their temporal order. Each vertex $v_k \in \mathcal{V}$ corresponds to a Label Function $L_{v_{k}} \in \mathcal{L}$, and an edge $(v_i, v_j) \in \mathcal{E}$ indicates temporal order between them. A path \( P = (v_1, v_2, \dots, v_n) \) is defined as a sequence of vertices such that each consecutive pair \( (v_i, v_{i+1}) \in \mathcal{E} \), with \( v_1 \) being a node of zero in-degree and \( v_n \) a node of zero out-degree. A Strategy Graph $\mathcal{G}$ contains multiple valid paths $P$ to capture diverse successful strategies for a given task.

\subsubsection{Trajectory Sampling and Categorization.}
\label{3.2.1}

In each iteration $i$, a set of trajectories $\{\tau_j\}$ is sampled from the policy that has been refined by our framework in the previous iteration, denoted as $\pi^{(i-1)}$. To promote diversity among trajectories, we introduce stochasticity into the action selection process by appropriately tuning specific parameters (details in Appendix~\ref{Parameters Configuration}).

After sampling, each trajectory $\tau$ is evaluated using the Strategy Graph $\mathcal{G}$. Formally, for a path $P$ in $\mathcal{G}$, we define the score of a sampled trajectory $\tau$ as a criterion for evaluation:
\begin{equation}
    \text{Score}_P(\tau) = \sum_{v \in P} L_v(\tau)
\end{equation}
Based on the score, we define a function $\mathcal{C}(\tau)$ that assigns each trajectory to one of the following categories:
\begin{equation}
    \mathcal{C}(\tau) =
    \begin{cases}
    \text{\emph{Fully Passed}}, & \text{if } \exists P \text{ s.t. } \text{Score}_P(\tau) = |P|, \\
    \text{\emph{Partially Passed}}, & \text{if } \exists P \text{ s.t. } 0 < \text{Score}_P(\tau) < |P|, \\
    \text{\emph{Failed}}, & \text{if } \forall P, \text{Score}_P(\tau) = 0.
    \end{cases}
\end{equation}
where \emph{Fully Passed} trajectories are directly incorporated into the training data, as they represent successful strategies. \emph{Partially Passed} trajectories are utilized for Graph Expansion (Section~\ref{3.2.2}) to enable strategic exploration. \emph{Failed} trajectories are leveraged in Failure-Driven Task Exploration (Section~\ref{3.3.2}) to improve generalization.

\subsubsection{Graph Expansion.}
\label{3.2.2}

The Strategy Graph $\mathcal{G}$ is initialized as a linear path derived from expert demonstrations. In each iteration $i$, we expand $\mathcal{G}^{(i)}$ by integrating novel strategies discovered in \emph{Partially Passed} trajectories $\tau$ that successfully complete the task as verified by environment feedback.

For each such trajectory $\tau$, we extract a new sequence of Label Functions $\mathcal{\tilde{L}}$ via Semantic Code Abstraction (Section~\ref{3.1}), thus forming a novel path $\tilde{P}_\tau = (\tilde{v}_1, \tilde{v}_2, \dots, \tilde{v}_n)$ where $L_{\tilde{v}_i} \in \mathcal{\tilde{L}}$. This new path $\tilde{P}_\tau$ is incorporated into the graph as an alternative valid path if it satisfies the goal condition. Specifically, we incorporate $\tilde{P}_\tau$ into $\mathcal{G}^{(i)}$ by merging identical vertices and extending the graph with new vertices and edges corresponding to novel steps. We formalize this process as:
\begin{equation}
    \mathcal{G}^{(i+1)} = \mathcal{G}^{(i)} \cup \left\{ \tilde{P}_\tau \;\middle|\; \mathcal{C}(\tau) = \text{\emph{Partially Passed}},\; \mathcal{F}(\tau) = 1 \right\},
\end{equation}
where the function $\mathcal{F}(\tau)$ returns $1$ if trajectory $\tau$ succeeds according to the environment feedback, and $0$ otherwise. 

The graph expansion mechanism enables Strategy Graph $\mathcal{G}$ to iteratively encode a richer strategy space. As $\mathcal{G}$ expands, it serves as a verifier to evaluate sampled trajectories and retain diverse and valid \textit{Fully Passed} trajectories, thereby enabling the agent to explore novel strategies beyond expert demonstrations and learn from a broader solution space, thus enhancing in-domain behavioral diversity.

\subsection{Trajectory-Guided Extrapolation}
\label{3.3}

To enhance out-of-domain diversity, Trajectory-Guided Extrapolation expands the task space beyond expert demonstrations by leveraging successful and failed trajectories.

\subsubsection{Success-Based Task Augmentation.}
\label{3.3.1}

While both Semantic Code Abstraction (Section~\ref{3.1}) and Strategy Graph Expansion (Section~\ref{3.2}) heavily rely on expert demonstrations, the set of tasks with such demonstrations is limited and static at initialization, restricting our framework's generalization.

To overcome this constraint and expand the set of tasks that can benefit from expert-like supervision, we deploy the current policy $\pi^{(i)}$ to conduct a thorough evaluation in the environment after each self-training iteration $i$. When $\pi^{(i)}$ successfully completes a new task not included in the original task pool, the corresponding trajectory \(\tau\) is regarded as a \textit{pseudo-expert demonstration}, and we add them into the task pool. Formally, the expansion process can be described as:
\begin{equation}
    \mathcal{T}^{(i+1)} = \mathcal{T}^{(i)} \cup \{ g_{\tau} \mid \mathcal{F}(\tau) = 1, g_{\tau} \notin \mathcal{T}^{(i)} \},
\end{equation}
where \(\mathcal{T}^{(i)}\) is the task pool at iteration \(i\), \(g_{\tau}\) is the task goal associated with the successful trajectory \(\tau\), and \(\mathcal{F}(\tau) = 1\) indicates that \(\tau\) successfully completes the task according to environment feedback.

This iterative process grows the supervised task space, enabling continual learning and improving generalization.

\subsubsection{Failure-Driven Task Exploration.}
\label{3.3.2}

In parallel, we utilize Failure-Driven Task Exploration to harness the potential hidden within failed trajectories. Although these trajectories fail under their original task intents, they may still represent valid and meaningful behaviors aligned with alternative task intents~\cite{andrychowicz2017hindsight}.

To unlock this potential, we analyze failed trajectories collected during Trajectory Sampling and infer plausible task intents using the model's reasoning capabilities. Since the initially inferred intents may be noisy or ambiguous, we further refine these intents to improve their quality. See the Appendix~\ref{Prompt for Failure-Driven Task Exploration} for detailed prompts. Formally, the process can be expressed as:
\begin{equation}
    g'_{\tau} = \mathcal{R}(\mathcal{I}(\tau)) ~ \text{s.t.} ~ \mathcal{C}(\tau) = \text{\emph{Failed}},
\end{equation}
where $\mathcal{I}(\tau)$ denotes the new task intent inferred from $\tau$, and $\mathcal{R}$ is the refinement function that improves the quality of the inferred intent to produce the final task intent $g'_{\tau}$.

These trajectories, along with their new aligned task intents, are then incorporated into the training data. This approach enables our framework to recycle failed experiences into useful training signals, enhancing data efficiency and enriching the diversity by introducing out-of-domain data.

\begin{table*}[!t]
\vspace{-10pt}
\centering
\renewcommand{\arraystretch}{1.2}   
\caption{Overall Performance and Generalization Performance on \emph{VisualWebArena} and \emph{AndroidWorld} with OS-Atlas.}
\label{main_result}
\setlength{\tabcolsep}{4pt} 

\begin{tabular}{ll cc cc cc cc cc}
\toprule
\multicolumn{2}{c}{\multirow{3}{*}{\textbf{Methods}}}
& \multicolumn{8}{c}{\textbf{VisualWebArena}} & \multicolumn{2}{c}{\multirow{2}{*}{\textbf{AndroidWorld}}} \\
\cmidrule(lr){3-10}
& & \multicolumn{2}{c}{\textbf{Classifieds}} & \multicolumn{2}{c}{\textbf{Reddit}} & \multicolumn{2}{c}{\textbf{Shopping}} & \multicolumn{2}{c}{\textbf{Average}} \\
\cmidrule(lr){3-4}\cmidrule(lr){5-6}\cmidrule(lr){7-8}\cmidrule(lr){9-10}\cmidrule(lr){11-12}
& & Overall & Gener. & Overall & Gener. & Overall & Gener. & Overall & Gener. & Overall & Gener. \\
\midrule

\multicolumn{2}{c}{Zero-shot}  & 3.42 & 4.23 & 0.95 & 1.59 & 1.29 & 1.43 & 1.76 & 2.19 & 0.00 & 0.00 \\
\midrule
\multicolumn{2}{c}{Baseline (BC)}  & 6.41 & 2.82 & 6.67 & 6.35 & 9.87 & 9.29 & 8.24 & 6.93 & 2.59 & 2.86 \\
\midrule

\multirow{3}{*}{DigiRL} 
& Iter.1 & 9.83 & 4.23 & 8.10 & 11.11 & 9.66 & 10.00 & 9.34 & 8.76 & 5.17 & 5.71 \\
& Iter.2 & 7.26 & 4.23 & 7.14 & 7.94 & 11.80 & 13.57 & 9.56 & 9.85 & 5.60 & 5.71 \\
& Iter.3 & 8.55 & 4.23 & 7.62 & 9.52 & 12.88 & 12.86 & 10.55 & 9.85 & 6.03 & 5.71 \\
\midrule

\multirow{3}{*}{BAGEL} 
& Iter.1 & 6.84 & 7.04 & 6.19 & 4.76 & 8.37 & 8.57 & 7.47 & 7.30 & 4.31 & 3.70 \\
& Iter.2 & 9.83 & 8.45 & 3.33 & 4.76 & 11.16 & 11.43 & 9.01 & 9.12 & 6.03 & 5.71 \\
& Iter.3 & 5.13 & 2.82 & 5.71 & 6.35 & 8.80 & 10.71 & 7.14 & 7.66 & 5.17 & 5.71 \\
\midrule

\multirow{3}{*}{Self-Improve} & Iter.1 & 7.26 & 2.82 & 7.14 & 7.94 & 8.15 & 7.86 & 7.69 & 6.57 & 7.76 & 5.71 \\
& Iter.2 & 6.84 & 5.63 & 5.71 & 4.76 & 8.15 & 8.57 & 7.25 & 6.93 & 7.33 & 11.43 \\
& Iter.3 & 7.26 & 2.82 & 3.81 & 4.76 & 9.87 & 10.71 & 7.80 & 7.30 & 9.05 & 11.43 \\
\midrule

\multirow{3}{*}{\textbf{CORE}} 
& Iter.1 & 11.11 & 5.63 & 7.62 & 9.52 & 12.02 & 7.14 & 10.77 & 7.30 & 10.34 & 8.57 \\
& Iter.2 & 11.54 & 8.45 & 8.10 & 11.11 & 12.45 & 7.14 & 11.21 & 8.39 & 12.07 & 11.43 \\
& Iter.3 & 11.97 & 14.08 & 7.14 & 7.94 & 13.73 & 11.43 & \textbf{11.76} & \textbf{11.31} & \textbf{14.22} & \textbf{14.29} \\
\bottomrule
\end{tabular}
\\ \footnotesize \vspace{2mm}
\textit{``Overall'' denotes all tasks; ``Gener.'' denotes only tasks from the generalization (test) split.}
\end{table*}

\section{Experiment}
\label{sec:experiment}

\subsection{Experimental Setup}

\subsubsection{Backbone.}
We evaluate our framework on \textbf{OS-Atlas-Base-7B}~\cite{wu2024atlas}, which is a GUI grounding model finetuned from Qwen2-VL-7B-Instruct~\cite{wang2024qwen2} and supports multimodal inputs. In our setting, we define a unified observation space across different environments (details in Appendix~\ref{Observation Space}).

\subsubsection{Benchmarks.}
We evaluate our framework in two interactive environments: \textbf{1) VisualWebArena}~\cite{koh2024visualwebarena}: A web-based benchmark for assessing agents' web navigation and interaction capabilities. \textbf{2) AndroidWorld}~\cite{rawles2024androidworld}: An Android-based benchmark for evaluating agents' operation of mobile apps.

\subsubsection{Compared Methods.}
To validate the effectiveness of our framework, we compare it against three representative approaches:
\begin{itemize}
    \item \textbf{Behavior Cloning (BC)}: We implement BC by training on expert demonstrations (details in Appendix~\ref{Expert Demonstrations Collection}). In our setting, BC serves as the baseline, providing an essential warm-up stage for stable iterative training due to the poor performance of zero-shot backbones.
    \item \textbf{Reinforcement Learning (RL)}: We compare with \textbf{DigiRL}~\cite{bai2024digirl}, an advanced RL-based framework that improves agent performance in a realistic Android-based environment through reward-driven exploration.
    \item \textbf{Self-Training}: We compare with two recent self-training approaches, \textbf{BAGEL}~\cite{murty2024bagel} and \textbf{Self-Improve}~\cite{patel2024large}, which refine agents through interaction with the environment under limited supervision.
\end{itemize}

\subsubsection{Implementation Details.}

\textbf{1)} To comprehensively evaluate agent performance, we randomly split each dataset into 70\% training and 30\% test sets (see Appendix~\ref{Train/Test Set Split} for details). The full dataset is used to assess overall performance, while the test set measures generalization performance. \textbf{2)} To demonstrate the iterative improvement capability of our framework, we conduct experiments with $N = 3$ self-training iterations. \textbf{3)} We use \textbf{Qwen2.5-Coder-3B-Instruct}~\cite{hui2024qwen2} as the code generation model for Label Function Synthesis; aside from this step, all self-training iterations rely solely on the agent's own model without using any external LLMs.

\subsection{Main Results}
We present the experimental results that demonstrate the effectiveness of the proposed CORE framework. Detailed quantitative results are summarized in Table~\ref{main_result}.

\paragraph{Superior Performance over Compared Methods.}
Compared to Behavior Cloning (baseline), RL-Based method DigiRL and two self-training methods BAGEL and Self-Improve, CORE achieves superior overall performance. Unlike the RL-based method DigiRL, our framework bypasses the need for handcrafted reward functions while still delivering stronger results. In comparison to other self-training approaches, CORE is significantly more cost-efficient, achieving a better performance-to-cost ratio; See Appendix~\ref{Exploration Cost Efficiency Analysis} for detailed cost efficiency analysis.

\begin{figure}[t]
\vspace{-5pt}
\centering
\includegraphics[width=0.47\textwidth]{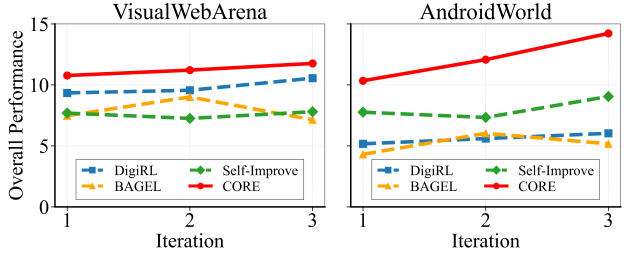}
\caption{Iterative performance comparison, highlighting CORE's steady and notable performance gains among three other methods.}
\label{fig:iterative_performance_comparison}
\end{figure}

\paragraph{Steady Performance Gains Across Iterations.}
As shown in Figure~\ref{fig:iterative_performance_comparison}, we find that CORE exhibits steady and notable performance improvements across iterations, consistently outperforming the other methods. This confirms the effectiveness of our iterative optimization strategy.

\paragraph{Robust Generalization to Unseen Tasks}
Our framework demonstrates remarkable generalization capabilities when applied to unseen tasks. Specifically, CORE achieves a substantial improvement in generalization scores, rising from 6.93 to \textbf{11.31} on \emph{VisualWebArena}, and from 2.86 to \textbf{14.29} on \emph{AndroidWorld}. These consistent and significant gains underscore the robustness of CORE's ability to generalize effectively to unseen scenarios.

\begin{table}[!t]
\centering
\caption{Ablation results of Strategy Graph Expansion (SGE) and Trajectory-Guided Extrapolation (TGE)}
\label{ablation}
\begin{tabular}{lcc|lcc}
\toprule
\multicolumn{3}{c|}{\textbf{VisualWebArena}} & \multicolumn{3}{c}{\textbf{AndroidWorld}} \\
\textbf{Methods} & Iter.1 & Iter.2 & \textbf{Methods} & Iter.1 & Iter.2 \\
\midrule
Baseline      & \multicolumn{2}{c|}{8.24}        & Baseline      & \multicolumn{2}{c}{2.59} \\
w/o SGE       & 9.45   & 10.44    & w/o SGE       & 6.03   & 6.90 \\
w/o TGE       & 9.45   & 10.66    & w/o TGE       & 5.17   & 5.17 \\
\textbf{CORE} & \textbf{10.77}  & \textbf{11.21} & \textbf{CORE} & \textbf{10.34} & \textbf{12.07} \\
\bottomrule
\end{tabular}
\end{table}

\subsection{Indepth Analysis.}

\paragraph{Ablation Study.}

We conduct ablation studies on two mechanisms: Strategy Graph Expansion (SGE) and Trajectory-Guided Extrapolation (TGE). Results in Table~\ref{ablation} show that removing either component leads to performance drops, especially in the second iteration, highlighting their synergistic effect in enhancing agent capability. Specifically, TGE shows greater impact on \emph{AndroidWorld}, which contains fewer tasks, possibly because task space expansion is more crucial when the dataset is limited. In contrast, SGE proves more critical on \emph{VisualWebArena}, which has a larger number of tasks, where enriching novel solutions within individual tasks becomes increasingly important as the overall task space saturates. Further ablation analysis of Success-Based Task Augmentation and Failure-Driven Task Exploration in TGE is provided in Appendix~\ref{Ablation Study on Trajectory-Guided Extrapolation}.

\begin{figure}[t]
\vspace{-5pt}
\centering
\includegraphics[width=0.47\textwidth]{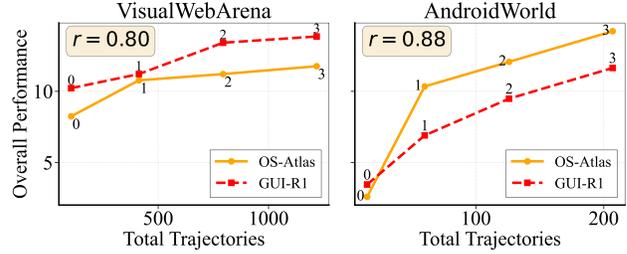}
\caption{Performance growth based on the number of training trajectories for two backbones on two benchmarks, showing a positive correlation with diminishing returns.}
\label{fig:performance_and_trajectory}
\end{figure}

\begin{table}[!t]
\centering
\renewcommand{\arraystretch}{1.3}
\caption{Overall Performance and Generalization Performance on \emph{VisualWebArena} and \emph{AndroidWorld} with GUI-R1.}
\label{main_result_GUI_R1}
\setlength{\tabcolsep}{3pt}
\begin{tabular}{c cc cc}
\toprule
\multirow{2}{*}{\textbf{Methods}} & \multicolumn{2}{c}{\textbf{VisualWebArena}} & \multicolumn{2}{c}{\textbf{AndroidWorld}} \\
\cmidrule(lr){2-3} \cmidrule(lr){4-5}
& Overall & Gener. & Overall & Gener. \\
\midrule
Zero-shot  & 0.33 & 0.00 & 1.72 & 2.86 \\
Baseline  & 10.22 & 8.76 & 3.45 & 2.86 \\
\midrule
DigiRL  & 10.11 & 8.76 & 6.90 & 8.57 \\
BAGEL  & 10.44 & 9.49 & 2.59 & 2.86 \\
Self-Improve  & 8.35 & 7.30 & 6.47 & 8.57 \\
\midrule
\textbf{CORE} & \textbf{11.21} & \textbf{9.85} & \textbf{6.90} & \textbf{8.57} \\
\bottomrule
\end{tabular}
\\ \footnotesize \vspace{2mm}
\textit{``Overall'' denotes all tasks; ``Gener.'' denotes only tasks from the generalization (test) split.}
\vspace{-10pt}
\end{table}

\paragraph{Model-Agnostic Capability.}
To further demonstrate that our framework is model-agnostic, we conduct additional experiments using another backbone, \textbf{GUI-R1-7B}~\cite{luo2025gui}, a GUI grounding model finetuned from Qwen2.5-VL-7B~\cite{bai2025qwen2}. As shown in Table~\ref{main_result_GUI_R1}, CORE consistently outperforms all compared methods on both benchmarks. These results clearly confirm that CORE is not limited to a specific model architecture, highlighting its general applicability across different backbones. More details and additional iteration results are provided in Appendix~\ref{Full Results on GUI-R1 Backbone}.

\paragraph{Performance vs. Trajectory Growth Analysis.}

We observe a strong positive correlation between the number of training trajectories and model performance on both \emph{VisualWebArena} and \emph{AndroidWorld} benchmarks, as shown in Figure~\ref{fig:performance_and_trajectory}. As trajectory volume grows through iterative self-training, the performance of OS-Atlas and GUI-R1 consistently improves, highlighting the robustness of our framework. However, diminishing returns are evident, particularly in \emph{AndroidWorld}, where the two backbones show significant gains in early iterations (0 to 1), but the performance curve flattens in later rounds, indicating smaller marginal gains with additional data beyond a certain point.

\begin{figure}[t]
\centering
\includegraphics[width=0.47\textwidth]{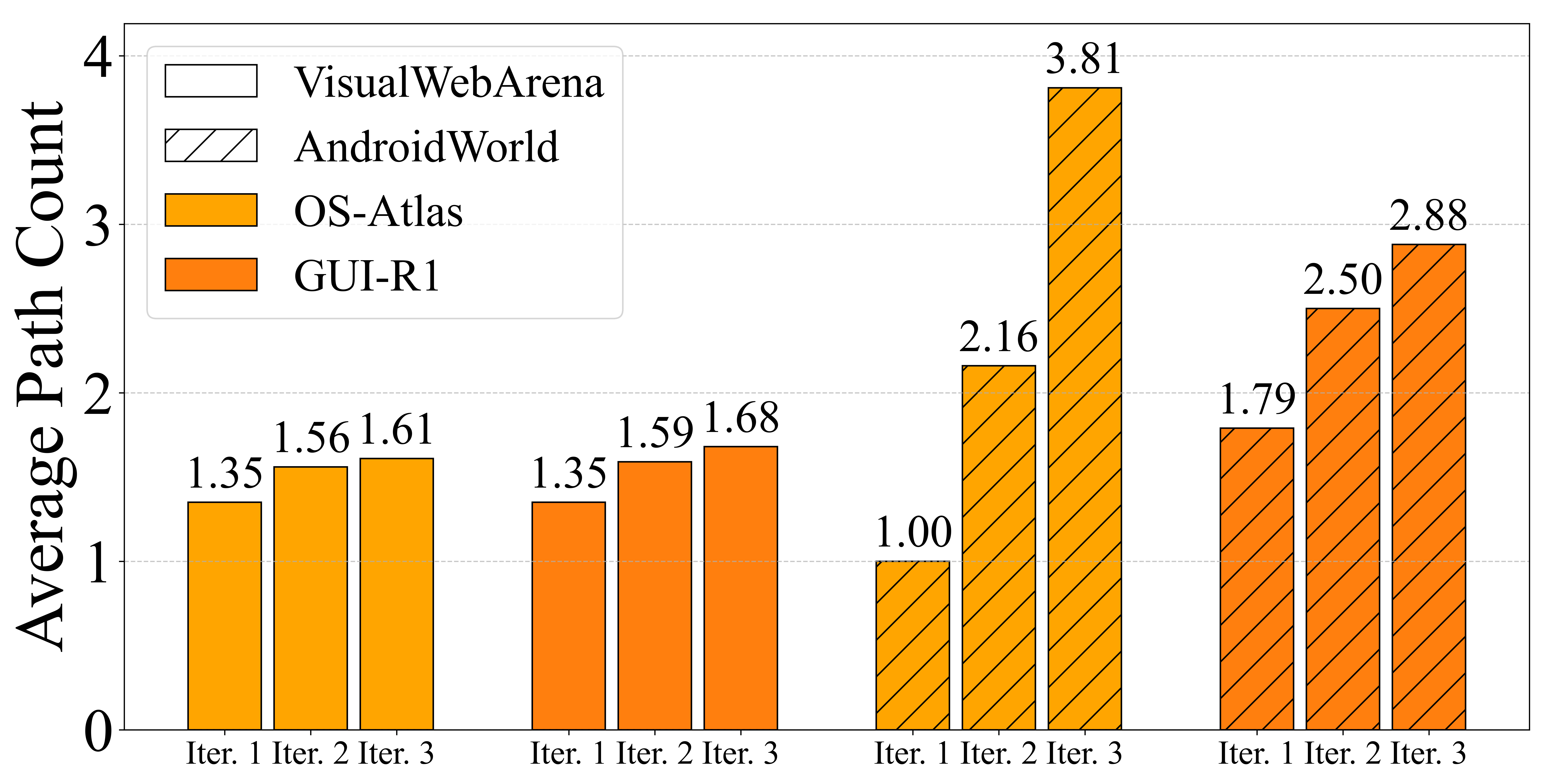}
\caption{Average path count in Strategy Graph over iterations on two benchmarks. All settings exhibit increased path count over iterations, validating the effectiveness of Strategy Graph Expansion in promoting behavioral diversity.}
\label{fig:path_count}
\end{figure}

\paragraph{Diversity in Strategy Graph.}

We evaluate the behavioral diversity induced by Strategy Graph Expansion using the average path count in Strategy Graph across iterations, as shown in Figure~\ref{fig:path_count}. On \emph{VisualWebArena}, both OS-Atlas and GUI-R1 show steady but moderate growth from 1.35 to 1.61 and 1.68, respectively, reflecting a gradual expansion of diverse behaviors. In contrast, on \emph{AndroidWorld}, OS-Atlas exhibits a sharp increase from 1.00 to 3.81, and GUI-R1 grows from 1.79 to 2.88, reflecting substantial gains in diversity. These results confirm that Strategy Graph effectively captures multiple valid paths and novel strategies, thereby enabling Strategy Graph Expansion to promote behavioral diversity. We further validate the effectiveness of Key Step Identification in Strategy Graphs and other Prompt-guided Step modules including Label Function Synthesis and New Intent Generation \& Refinement; see Appendix~\ref{Prompt-guided Step Evaluations} for detailed evaluations.

\paragraph{Case Study.}

\begin{figure}[t]
\centering
\includegraphics[width=0.45\textwidth]{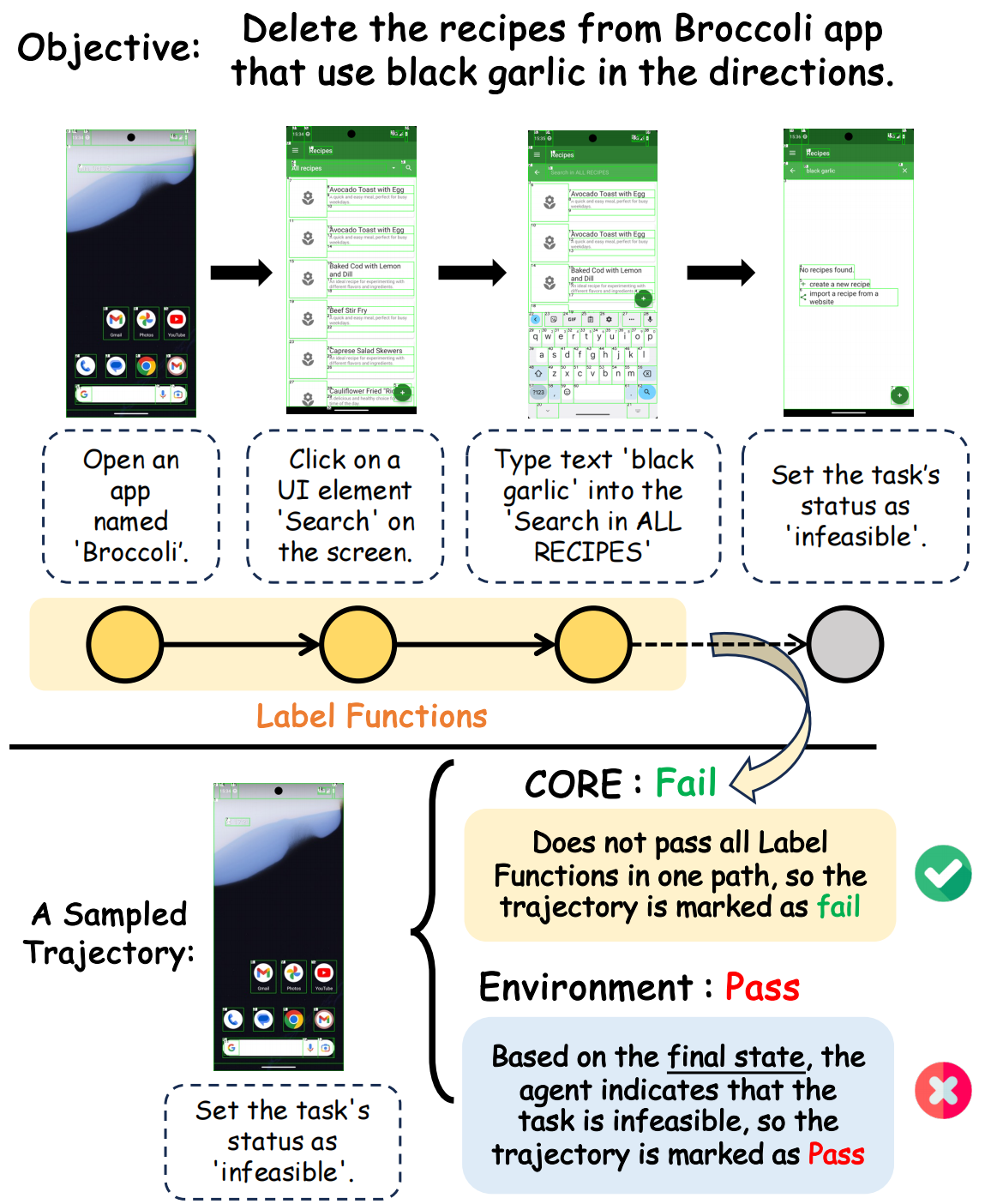}
\caption{An example demonstrating how Label Functions validate key steps and prevent misjudgments inherent in result-based benchmarks that evaluate only the final state.}
\label{fig:experiment_example}
\end{figure}

To illustrate the importance of Label Functions, we visualize a case, as shown in Figure~\ref{fig:experiment_example}. This task is infeasible because no such recipes exist in the app. The result-based benchmark \emph{AndroidWorld} erroneously pass an agent that simply declares the task infeasible without necessary actions. In contrast, our Label Functions verify that the agent opens the Broccoli app, searches for ``black garlic'' and confirms the absence of matching recipes. This key-process-based evaluation ensures that the trajectory is correctly assessed, thus ensuring a more reliable and robust evaluation. See Appendix~\ref{CORE vs. Environment-Only Learning} for quantitative results comparing CORE with the method that relies solely on environment feedback, and Appendix~\ref{Case Study of Label Functions} for more case studies of Label Functions.

\section{Conclusion}
\label{sec:conclusion}
In conclusion, we introduce \textbf{CORE}, a Code-based Inverse Self-Training Framework with Graph Expansion that bridges the gap between imitation and exploration for training multimodal virtual agents. By leveraging \textbf{Semantic Code Abstraction}, CORE infers key process reward functions from expert demonstrations without manual design. The \textbf{Strategy Graph Expansion} and \textbf{Trajectory-Guided Extrapolation} modules substantially enhance in-domain and out-of-domain behavioral diversity respectively. Extensive experiments conducted across Web and Android environments show that CORE significantly boosts overall performance and generalization, outperforming existing methods and achieving steady gains across successive iterations, validating its effectiveness for robust virtual agent training.
\newpage
{
    \small
    \bibliographystyle{ieeenat_fullname}
    \bibliography{main}
}

\clearpage
\setcounter{page}{1}
\maketitlesupplementary

\appendix

\lstset{
    literate={“}{{\textquotedblleft}}1
             {”}{{\textquotedblright}}1
             {‘}{{\textquoteleft}}1
             {'}{{\textquoteright}}1,
    breaklines=true, 
}

\section*{Overview}
In this supplementary material we present:
\begin{itemize}[itemsep=1pt]
    \item Detailed prompts for Key Step Identification, Label Function Synthesis, and Failure-Driven Task Exploration are provided in Section~\ref{Prompt Design Details}.
    \item The experimental setup is described in detail in Section~\ref{Experimental Setup Details}.
    \item Pseudocode for our three methods and the overall framework pipeline is provided in Section~\ref{Pseudocode}.
    \item More analysis and discussion are included in Section~\ref{Analysis and Discussion}.
    \item Prompt-guided step evaluations are included in Section~\ref{Prompt-guided Step Evaluations}.
    \item More case studies are presented in Section~\ref{Case Studies}.
\end{itemize}

\section{Prompt Design Details}
\label{Prompt Design Details}
This section offers a detailed summary of the prompt designs employed in our experiments.

\subsection{Prompt for Key Step Identification}
\label{Prompt for Key Step Identification}

This subsection outlines the prompt used to identify key steps from actions' semantic descriptions in a given trajectory.

\begin{tcolorbox}[
    colback=gray!10!white,
    colframe=gray!50!black,
    title=Prompt for Key Step Identification,
    fonttitle=\bfseries,
    boxrule=0.5mm,
    arc=2mm,
    width=\columnwidth,
    breakable,
    before skip=2mm,
    after skip=2mm,
    left=3pt,
    right=3pt,
    top=3pt,
    bottom=3pt
]
{\ttfamily\small
\textbf{You will be provided with:} \\
- \textbf{Objective}: A clear task or question to be addressed. \\
- \textbf{Successful Action Sequence}: A set of steps demonstrating how the \textbf{Objective} was previously achieved. \\

\textbf{Your task:} \\
Analyze the provided "Objective" and the "Successful Action Sequence" to extract the \textbf{key action sequence}. The key action sequence should include only the essential steps that directly contribute to achieving the Objective. \\
- Exclude irrelevant, redundant, or non-critical actions from the \textbf{Successful Action Sequence}. Focus solely on the necessary steps required for success, based on your understanding of the Objective and any contextual information from the images (if provided). But Do not modify the original wording of any action. \\
- Do not simply repeat and output the \textbf{Successful Action Sequence}; critically evaluate and distill it into a concise list of core actions. \\

\textbf{Output:} \\
Present the key action sequence as a numbered list without any additional explanation outside the list: \\
1. Key action 1 \\
2. Key action 2 \\
...
}
\end{tcolorbox}

\subsection{Prompt for Label Function Synthesis}
\label{Prompt for Label Function Synthesis}

This subsection describes the prompt designed for synthesizing Label Functions based on key steps, where the $<<$API\_FUNCTIONS$>>$ and $<<$GUIDANCE$>>$ vary by benchmark

\lstset{
    basicstyle=\ttfamily\small,
    breaklines=true,
    breakatwhitespace=true,
    tabsize=2,
    showspaces=false,
    showstringspaces=false,
    keepspaces=true,
    numbers=none,
    columns=fullflexible,
    breakindent=0pt,
    gobble=0,
    linewidth=\linewidth,
    xleftmargin=0pt,
    xrightmargin=0pt
}

\begin{tcolorbox}[
    colback=gray!10!white,
    colframe=gray!50!black,
    title=Prompt for Label Function Synthesis,
    fonttitle=\bfseries,
    boxrule=0.5mm,
    arc=2mm,
    width=\columnwidth,
    breakable,
    before skip=2mm,
    after skip=2mm,
    left=3pt,
    right=3pt,
    top=3pt,
    bottom=3pt
]
{\ttfamily\small
You are an AI assistant tasked with using the correct API functions to generate a verification function in Python to determine whether a specific task has been completed based on a Trajectory. The task is a fixed text description provided below, representing the current goal that the function should evaluate. The Trajectory is a sequence of actions represented as strings, which you will analyze using the provided API functions. \\

Here are the available API functions you can use: \\
<<API\_FUNCTIONS>> \\

You will be provided with the following code format: \\

```python
\begin{lstlisting}
from Function_APIs import *
def verify_function(trajectory):
    # Check each API call condition sequentially
    # Only one <API_call> in one if statement, avoiding two <API_call> with 'and' or nested if statements.
    if not <API_call_1>:
        return False
    if not <API_call_2>:
        return False
    # Add more <API_call> if needed
    # Return True only if all conditions were satisfied
    return True
\end{lstlisting}
``` \\

The verification function should: \\
1. Use the provided API functions to validate actions within the Trajectory. \\
2. Simply output the code only without explanation and example usage. \\
3. Each if statement can only contain one API call, and cannot have nested if statements. \\

Guidance: \\
<<GUIDANCE>>
}
\end{tcolorbox}

\subsection{Prompt for Failure-Driven Task Exploration}
\label{Prompt for Failure-Driven Task Exploration}

This subsection introduces the prompts used for Failure-Driven Task Exploration, including the prompt for generating new intent and prompt for refining intent, where the $<<$EXAMPLES$>>$ used in the refinement prompt vary across benchmarks.

\begin{tcolorbox}[
    colback=gray!10!white,
    colframe=gray!50!black,
    title=Prompt for New Intent Generation,
    fonttitle=\bfseries,
    boxrule=0.5mm,
    arc=2mm,
    width=\columnwidth,
    breakable,
    before skip=2mm,
    after skip=2mm,
    left=3pt,
    right=3pt,
    top=3pt,
    bottom=3pt
]
{\ttfamily\small
Below is a trajectory to complete a task. Please Analyze the trajectory and write a reasonable task intent that the trajectory does successfully complete.
}
\end{tcolorbox}

\begin{tcolorbox}[
    colback=gray!10!white,
    colframe=gray!50!black,
    title=Prompt for Intent Refinement,
    fonttitle=\bfseries,
    boxrule=0.5mm,
    arc=2mm,
    width=\columnwidth,
    breakable,
    before skip=2mm,
    after skip=2mm,
    left=3pt,
    right=3pt,
    top=3pt,
    bottom=3pt
]
{\ttfamily\small
You are given a candidate "task intent" that was automatically generated from a trajectory. Your job is to transform it into a clean, imperative-style intent, leave it unchanged if it already meets the criteria, or flag it as INVALID if it doesn't describe a real intent. Follow these rules exactly: \\

1. \textbf{Preserve well-formed intents}: If the candidate already adheres to all the rules below (clean, imperative style with a clear action and explicit object), return it exactly as is. \\
2. \textbf{No prefixes or labels}: Don't start with "The task is," "New task intent:," "This intent," or any extra words. \\
3. \textbf{Validity check}: If it doesn't describe a real intent or is meaningless, respond with exactly `INVALID`. For example, "New task intent:", "OBSERVATION: ", or any other non-intent placeholder should be flagged as `INVALID`. \\
4. \textbf{Explicit object requirement}: The intent must include both a clear action *and* an explicit object or target. If the candidate has no specified object (e.g. "Add to cart", "Compare the prices of the products", "Stop", "Go back"), respond with `INVALID`. \\
5. \textbf{Prohibit negation or interruption intents}: If the candidate describes stopping, canceling, preventing, or otherwise negating another action (e.g., "The task intent is to stop the..."), respond with `INVALID`. \\

Examples:\\
<<EXAMPLES>>
}
\end{tcolorbox}

\section{Experimental Setup Details}
\label{Experimental Setup Details}
\subsection{Observation Space}
\label{Observation Space}

To ensure a fair comparison and eliminate the influence of different observation modalities across environments, we adopt a unified observation space for all agents in both web-based and Android-based settings. It consists of three components: 
\begin{itemize}
    \item \textbf{Screenshot with SoM}: A visual representation of the current GUI state, augmented with Set-of-Marks (SoM) annotations that highlight clickable or interactive elements, enabling the model to better focus on key regions.
    \item \textbf{SoM Elements}: A structured metadata list corresponding to the SoM-annotated elements, containing attributes such as text content and bounding boxes, thereby offering precise semantic grounding for potential actions.
    \item \textbf{Natural Language Instructions}: Task-specific instructions provided in natural language format, guiding the agent toward intended high-level goals within the current task context.
\end{itemize}

\subsection{Expert Demonstrations Collection}
\label{Expert Demonstrations Collection}

\begin{table}[ht]
\renewcommand{\arraystretch}{1.2}
\centering
\caption{The number of expert demonstrations by category for \textit{VisualWebArena} and \textit{AndroidWorld}}
\label{expert_demo}
\begin{tabular}{clc} 
\toprule
\textbf{Model \& Bench} & \textbf{Category} & \textbf{Count} \\
\midrule
    \multirow{4}{*}{\makecell{GPT-4V \\ on \\ \textit{VisualWebarena}}} 
    & Classifieds & \textbf{24} / 234 \\
    & Reddit      & \textbf{36} / 210 \\
    & Shopping    & \textbf{90} / 466 \\
    \cmidrule(lr){2-3}
    & Overall     & \textbf{150} / 910 \\
\cmidrule(lr){1-3}
    \makecell{Qwen2.5-VL-32B \\ on \\ \textit{Android World}} & Overall & \textbf{22} / 116 \\
\bottomrule
\end{tabular}
\\ \footnotesize \vspace{2mm}
\textit{Note: Each value indicates the number of expert demonstrations/ the total number of tasks.}
\end{table}
Expert demonstrations are successful trajectories executed by humans or external powerful models. Due to limited access to human-collected demonstrations, we instead relied on powerful external models to generate expert demonstrations. To collect these, we ran the entire benchmark in each environment using these models and extracted only the successful trajectories. The details of expert demonstrations in each environment are summarized in Table~\ref{expert_demo}. \textbf{1)} For \emph{VisualWebArena}, a total of \textbf{150} expert demonstrations were extracted from the officially provided evaluation set, using \textbf{GPT-4V} as the expert model. These demonstrations span three subsets: 24 in Classifieds, 36 in Reddit, and 90 in Shopping, covering approximately \textbf{16.5\%} of all tasks. \textbf{2)} For \emph{AndroidWorld}, we utilized the \textbf{Qwen2.5-VL-32B-Instruct} model to generate \textbf{22} expert demonstrations out of 116 tasks, resulting in a \textbf{19\%} coverage. These high-quality expert trajectories, collected from external powerful models, not only provided reliable supervision for Behavior Cloning but also laid the groundwork for our entire approach. They serve as the foundation for both the initial policy learning and our iterative self-training pipeline.

\subsection{Train/Test Set Split}
\label{Train/Test Set Split}

\begin{table}[ht]
\renewcommand{\arraystretch}{1.2}
\centering
\caption{Test set split by category for VisualWebArena and Android World}
\label{train_test_split}
\begin{tabular}{clc} 
\toprule
\textbf{Bench} & \textbf{Category} & \textbf{Test Set Split} \\
\midrule
    \multirow{4}{*}{\textit{VisualWebarena}} 
    & Easy &  62 / 205 (30.2\%) \\
    & Medium      & 114 / 378 (30.2\%) \\
    & Hard    & 98 / 327 (30.0\%) \\
    \cmidrule(lr){2-3}
    & Overall     & 274 / 910 (30.1\%)\\
\cmidrule(lr){1-3}
    \textit{Android World} & Overall & 35 / 116 (30.2\%) \\
\bottomrule
\end{tabular}
\\ \footnotesize \vspace{2mm}
\textit{Note: Each value indicates the number of test set / the total number of tasks.}
\end{table}

To evaluate both the overall performance and generalization of our iterative self-training framework on unseen tasks, we split the VisualWebArena and AndroidWorld datasets into training and test sets with a 7:3 ratio. \textbf{1)} For \emph{VisualWebArena}, we stratify the split to ensure Easy, Medium, and Hard difficulty levels each follow the ratio, preserving task diversity. \textbf{2)} For \emph{AndroidWorld}, we simply apply a totally random split. These test sets enable robust evaluation of the agents' generalization on unseen tasks across different environments. Details in Table~\ref{train_test_split}.

\subsection{Parameters Configuration}
\label{Parameters Configuration}

\textbf{1) Sampling stage:} To encourage the diversity of collected trajectories, we configure the sampling parameters as follows: $temperature = 1.0$, $top\_p = 0.9$, $top\_k = 50$, and $do\_sample = True$. These settings promote exploration of varied action sequences, enabling the agent to discover novel strategies. Furthermore, we set the number of sampled trajectories per task to 5 to ensure sufficient data for iterative self-training. \textbf{2) Training stage:} We employ the Llama-Factory framework for model training. To ensure stable convergence, we typically train for 10 epochs. However, in scenarios with extremely limited training data, we reduce the number of epochs to 5 to mitigate the risk of overfitting and maintain generalization performance.

\begin{algorithm}[t!]
    \caption{Semantic Code Abstraction}
    \label{alg:semantic_code_abstraction}
    \scriptsize
\begin{algorithmic}[1]
    \STATE \textbf{Input:} Trajectory $\tau = \{(s_1, a_1), \dots, (s_T, a_T)\}$, task goal $g$, API set $\mathcal{A}$.
    \STATE \textbf{Output:} A set of Label Functions $\mathcal{L}$.
    \STATE
    \STATE \textit{// Semantic Description Extraction}
    \STATE Initialize semantic description set $\mathcal{D}(\tau) \leftarrow \emptyset$.
    \FOR{each state-action pair $(s_t, a_t)$ in $\tau$}
        \STATE Map to semantic description: $d_t \leftarrow \mathcal{D}(s_t, a_t)$.
        \STATE $\mathcal{D}(\tau) \leftarrow \mathcal{D}(\tau) \cup \{d_t\}$.
    \ENDFOR
    \STATE
    \STATE \textit{// Key Step Identification}
    \STATE Identify key steps: $\mathcal{D}_{\text{key}} \leftarrow \mathcal{K}(\mathcal{D}(\tau), g)$.
    \STATE
    \STATE \textit{// Label Function Synthesis}
    \STATE Initialize Label Function set $\mathcal{L} \leftarrow \emptyset$.
    \FOR{each key step description $d_k$ in $\mathcal{D}_{\text{key}}$}
        \STATE Synthesize Label Function: $L_k \leftarrow \mathcal{S}(d_k, \mathcal{A})$.
        \STATE $\mathcal{L} \leftarrow \mathcal{L} \cup \{L_k\}$.
    \ENDFOR
    \STATE
    \STATE \textbf{return} $\mathcal{L}$
\end{algorithmic}
\end{algorithm}

\begin{algorithm}[t!]
    \caption{Strategy Graph Expansion}
    \label{alg:strategy_graph_expansion}
    \scriptsize
\begin{algorithmic}[1]
    \STATE \textbf{Input:} Sampled trajectories $\{\tau\}$, current Strategy Graph $\mathcal{G}^{(i)}$.
    \STATE \textbf{Output:} Updated graph $\mathcal{G}^{(i+1)}$, augmented data $\mathcal{D}_{\text{aug}}$, failed trajectories $\mathcal{D}_{\text{failed}}$.
    \STATE
    \STATE \textit{// Trajectory Categorization before Expansion}
    \STATE Initialize sets: $\mathcal{D}_{\text{partially}} \leftarrow \emptyset$.
    \FOR{each trajectory $\tau$ in $\{\tau\}$}
        \STATE Categorize $\tau$ using $\mathcal{G}^{(i)}$: $category \leftarrow \mathcal{C}(\tau)$.
        \IF{$category = \text{\emph{Partially Passed}}$}
            \STATE $\mathcal{D}_{\text{partially}} \leftarrow \mathcal{D}_{\text{partially}} \cup \{\tau\}$.
        \ENDIF
    \ENDFOR
    \STATE
    \STATE \textit{// Graph Expansion}
    \STATE Initialize updated graph $\mathcal{G}^{(i+1)} \leftarrow \mathcal{G}^{(i)}$.
    \FOR{each partially passed trajectory $\tau$ in $\mathcal{D}_{\text{partially}}$}
        \IF{task is successful based on environment feedback $\mathcal{F}(\tau) = 1$}
            \STATE Generate new label functions $\mathcal{\tilde{L}}_\tau$ from $\tau$ using Algorithm~\ref{alg:semantic_code_abstraction}.
            \STATE Form new path $\tilde{P}_\tau$ from $\mathcal{\tilde{L}}_\tau$.
            \STATE Expand graph: $\mathcal{G}^{(i+1)} \leftarrow \mathcal{G}^{(i+1)} \cup \{ \tilde{P}_\tau \}$.
        \ENDIF
    \ENDFOR
    \STATE
    \STATE \textit{// Trajectory Categorization after Expansion}
    \STATE Initialize sets: $\mathcal{D}_{\text{fully}} \leftarrow \emptyset$, $\mathcal{D}_{\text{failed}} \leftarrow \emptyset$.
    \FOR{each trajectory $\tau$ in $\{\tau\}$}
        \STATE Categorize $\tau$ using $\mathcal{G}^{(i+1)}$: $category \leftarrow \mathcal{C}(\tau)$.
        \IF{$category = \text{\emph{Fully Passed}}$}
            \STATE $\mathcal{D}_{\text{fully}} \leftarrow \mathcal{D}_{\text{fully}} \cup \{\tau\}$.
        \ELSIF{$category = \text{\emph{Failed}}$}
            \STATE $\mathcal{D}_{\text{failed}} \leftarrow \mathcal{D}_{\text{failed}} \cup \{\tau\}$.
        \ENDIF
    \ENDFOR
    \STATE Augment training data: $\mathcal{D}_{\text{aug}} \leftarrow \mathcal{D}_{\text{fully}}$.
    \STATE \textbf{return} $\mathcal{G}^{(i+1)}, \mathcal{D}_{\text{aug}}, \mathcal{D}_{\text{failed}}$
\end{algorithmic}
\end{algorithm}

\section{Pseudocode}
\label{Pseudocode}
\subsection{Semantic Code Abstraction}
This algorithm (as shown in Algorithm~\ref{alg:semantic_code_abstraction}) aims to convert a raw trajectory that contains a sequence of states and actions into a higher-level, structured representation. By identifying the key steps within a task and synthesizing them into a set of Label Functions, it establishes a semantic foundation for subsequent policy learning and generalization.

\subsection{Strategy Graph Expansion}
This algorithm (as shown in Algorithm~\ref{alg:strategy_graph_expansion}) details the dynamic expansion process of the Strategy Graph. It takes trajectories generated by the current policy and identifies those that represent new and successful solutions for solving a task. By leveraging Semantic Code Abstraction, these new strategies are integrated as new paths into the graph, thereby continuously enriching the agent's strategic knowledge base.

\subsection{Trajectory-Guided Extrapolation}
This algorithm (as shown in Algorithm~\ref{alg:trajectory_guided_extrapolation}) introduces a mechanism for task exploration and data augmentation. It expands the pool of known tasks by verifying new successes, but more critically, it extrapolates from failures. By inferring a new, plausible task intent that a failed trajectory might have inadvertently accomplished, the algorithm creatively transforms failed experiences into valuable training data, driving the agent to explore a wider range of tasks.

\setlength{\textfloatsep}{10pt}
\begin{algorithm}[t!]
    \caption{Trajectory-Guided Extrapolation}
    \label{alg:trajectory_guided_extrapolation}
    \scriptsize
\begin{algorithmic}[1]
    \STATE \textbf{Input:} Current task pool $\mathcal{T}^{(i)}$, trajectories set $E$ evaluated by $\pi$, \text{\emph{Failed}} trajectories set $\mathcal{D}_{\text{failed}}$.
    
    \STATE \textbf{Output:} Updated task pool $\mathcal{T}^{(i+1)}$, augmented data $\mathcal{D}_{\text{aug}}$.
    \STATE
    \STATE Initialize $\mathcal{T}_{\text{new}} \leftarrow \mathcal{T}^{(i)}$.
    \STATE
    \STATE \textit{// Success-Based Task Augmentation}
    \FOR{each trajectory $\tau$ in $E$}
        \IF{task succeeded $\mathcal{F}(\tau) = 1$ and its goal $g_\tau \notin \mathcal{T}^{(i)}$}
            \STATE Add new task to pool: $\mathcal{T}_{\text{new}} \leftarrow \mathcal{T}_{\text{new}} \cup \{g_\tau\}$.
        \ENDIF
    \ENDFOR
    \STATE Update task pool $\mathcal{T}^{(i+1)} \leftarrow \mathcal{T}^{(i)} \cup \mathcal{T}_{\text{new}}$
    \STATE
    \STATE \textit{// Failure-Driven Task Exploration}
    \STATE Initialize sets: $\mathcal{D}_{\text{aug}} \leftarrow \emptyset$.
    \FOR{each trajectory $\tau$ in $\mathcal{D}_{\text{failed}}$}
        \STATE Infer new plausible task intent: $g'_{\text{inferred}} \leftarrow \mathcal{I}(\tau)$.
        \STATE Refine the inferred intent for quality: $g'_{\tau} \leftarrow \mathcal{R}(g'_{\text{inferred}})$.
        \STATE Add training data: $\mathcal{D}_{\text{aug}} \leftarrow \mathcal{D}_{\text{aug}} \cup \{ (\tau, g'_\tau) \}$.
    \ENDFOR
    
    \STATE
    \STATE \textbf{return} $\mathcal{T}^{(i+1)}, \mathcal{D}_{\text{aug}}$
\end{algorithmic}
\end{algorithm}

\subsection{Overall Iterative Self-Training Pipeline}
This is the overall pipeline of our framework (as shown in Algorithm~\ref{alg:core_pipeline}), outlining an iterative self-training loop. After an initial setup with expert data, the pipeline repeatedly executes a cycle of Strategy Graph Expansion, Trajectory-Guided Extrapolation and data aggregation for a policy update. By continuously discovering new strategies and exploring new tasks in each iteration, the agent's policy is progressively enhanced.

\begin{algorithm}[t!]
    \caption{Overall Iterative Self-Training Pipeline}
    \label{alg:core_pipeline}
    \scriptsize
\begin{algorithmic}[1]
    \STATE \textbf{Input:} Expert demonstrations $\mathcal{D}_{\text{expert}}$, initial policy $\pi^{(0)}$, number of iterations $N$.
    \STATE \textbf{Output:} Final policy $\pi^{(N)}$.
    \STATE
    \STATE \textit{// Initialization (i=0)}
    \STATE Initialize training data $\mathcal{D}_{\text{train}}^{(0)} \leftarrow \mathcal{D}_{\text{expert}}$.
    \STATE Initialize task pool $\mathcal{T}^{(0)}$ with goals from $\mathcal{D}_{\text{expert}}$.
    \STATE Initialize Strategy Graph $\mathcal{G}^{(0)}$ by applying Algorithm~\ref{alg:semantic_code_abstraction} to each $\tau \in \mathcal{D}_{\text{expert}}$ to form initial paths.
    \STATE Fine-tune baseline policy: $\pi^{(0)} \leftarrow \text{FineTune}(\pi_{\text{base}}, \mathcal{D}_{\text{train}}^{(0)})$.
    \STATE
    \FOR{iteration $i = 0, \dots, N-1$}
        \STATE \textit{// 1. Trajectory Sampling}
        \STATE Sample a set of trajectories $\{\tau\}$ by executing $\pi^{(i)}$ on tasks from $\mathcal{T}^{(i)}$.
        \STATE
        \STATE \textit{// 2. Strategy Graph Expansion (Section~\ref{3.2})}
        \STATE $(\mathcal{G}^{(i+1)}, \mathcal{D}_{\text{aug\_1}}, \mathcal{D}_{\text{failed}}) \leftarrow \text{StrategyGraphExpansion}(\{\tau\}, \mathcal{G}^{(i)})$.
        \STATE
        \STATE \textit{// 3. Trajectory-Guided Extrapolation (Section~\ref{3.3})}
        \STATE Evaluate a set of trajectories $E$ by executing $\pi^{(i)}$ on whole benchmark.
        \STATE $(\mathcal{T}^{(i+1)}, \mathcal{D}_{\text{aug\_2}}) \leftarrow \text{TrajectoryGuidedExtrapolation}(\mathcal{T}^{(i)}, E, \mathcal{D}_{\text{failed}})$.
        \STATE
        \STATE \textit{// 4. Data Aggregation and Policy Update}
        \STATE Update training data: $\mathcal{D}_{\text{train}}^{(i+1)} \leftarrow \mathcal{D}_{\text{train}}^{(i)} \cup \mathcal{D}_{\text{aug\_1}} \cup \mathcal{D}_{\text{aug\_2}}$.
        \STATE Update policy: $\pi^{(i+1)} \leftarrow \text{FineTune}(\pi^{(i)}, \mathcal{D}_{\text{train}}^{(i+1)})$.
    \ENDFOR
    \STATE
    \STATE \textbf{return} $\pi^{(N)}$
\end{algorithmic}
\end{algorithm}

\section{Analysis and Discussion}
\label{Analysis and Discussion}

\subsection{Full Results on GUI-R1 Backbone.}
\label{Full Results on GUI-R1 Backbone}
We provide the full experimental results obtained on the GUI-R1-7B backbone in Table~\ref{all_result_GUI_R1}. These detailed results include per-domain performance on \emph{VisualWebArena} (Classifieds, Reddit, Shopping) and the CORE's complete evaluation across all iterations. The consistent improvements from Iteration 1 to Iteration 3 demonstrate that our framework maintains stable and progressive performance gains even on a different backbone, further validating its model-agnostic nature and robustness across multimodal architectures.

\begin{table*}[!h]
\centering
\renewcommand{\arraystretch}{1.3}
\caption{Overall Performance and Generalization Performance on \emph{VisualWebArena} and \emph{AndroidWorld} with GUI-R1.}
\label{all_result_GUI_R1}
\setlength{\tabcolsep}{3pt}

\begin{tabular}{ll cc cc cc cc cc}
\toprule
\multicolumn{2}{c}{\multirow{3}{*}{\textbf{Methods}}}
& \multicolumn{8}{c}{\textbf{VisualWebArena}} & \multicolumn{2}{c}{\multirow{2}{*}{\textbf{AndroidWorld}}} \\
\cmidrule(lr){3-10}
& & \multicolumn{2}{c}{\textbf{Classifieds}} & \multicolumn{2}{c}{\textbf{Reddit}} & \multicolumn{2}{c}{\textbf{Shopping}} & \multicolumn{2}{c}{\textbf{Average}} \\
\cmidrule(lr){3-4}\cmidrule(lr){5-6}\cmidrule(lr){7-8}\cmidrule(lr){9-10}\cmidrule(lr){11-12}
& & Overall & Gener. & Overall & Gener. & Overall & Gener. & Overall & Gener. & Overall & Gener. \\
\midrule
\multicolumn{2}{c}{Zero-shot}  & 0.00 & 0.00 & 0.48 & 0.00 & 0.43 & 0.00 & 0.33 & 0.00 & 1.72 & 2.86 \\
\multicolumn{2}{c}{Baseline}  & 12.82 & 11.27 & 7.62 & 6.35 & 10.09 & 8.57 & 10.22 & 8.76 & 3.45 & 2.86 \\
\midrule
\multicolumn{2}{c}{BAGEL}  & 11.11 & 11.27 & 7.62 & 6.35 & 11.37 & 10.00 & 10.44 & 9.49 & 2.59 & 2.86 \\
\multicolumn{2}{c}{Self-Improve}  & 12.39 & 12.68 & 6.19 & 4.76 & 7.30 & 5.71 & 8.35 & 7.30 & 6.47 & 8.57 \\
\multicolumn{2}{c}{DigiRL}  & 12.39 & 12.68 & 7.14 & 6.35 & 10.30 & 7.86 & 10.11 & 8.76 & 6.90 & 8.57 \\
\midrule
\multirow{3}{*}{\textbf{CORE}} 
& Iter.1 & 12.82 & 11.27 & 8.10 & 6.35 & 11.80 & 10.71 & 11.21 & 9.85 & 6.90 & 8.57 \\
& Iter.2 & 16.24 & 14.08 & 9.05 & 4.76 & 13.95 & 12.14 & 13.41 & \textbf{10.95} & 9.48 & 8.57 \\
& Iter.3 & 17.95 & 14.08 & 10.00 & 9.52 & 13.52 & 7.86 & \textbf{13.85} & 9.85 & \textbf{11.64} & \textbf{11.43} \\
\bottomrule
\end{tabular}
\\ \footnotesize \vspace{2mm}
\textit{``Overall'' denotes all tasks; ``Gener.'' denotes only tasks from the generalization (test) split.}
\end{table*}

\begin{table}[!t]
\centering
\caption{Ablation Study of the Success-Based Task Augmentation (SBTA) and Failure-Driven Task Exploration (FDTE) components within the Trajectory-Guided Extrapolation (TGE).}
\label{tab:ablation_study1}
\begin{tabular}{lcc}
\toprule
    \textbf{Method} & \textbf{Iter 1} & \textbf{Iter 2} \\
\midrule
    Baseline & \multicolumn{2}{c}{2.59} \\
\midrule
    w/o TGE & 5.17 & 5.17 \\
    w/o SBTA in TGE & 6.90 & 7.76 \\
    w/o FDTE in TGE & 7.76 & 9.48 \\
\midrule
    \textbf{CORE} & \textbf{10.34} & \textbf{12.07} \\
\bottomrule
\end{tabular}
\end{table}

\subsection{Ablation Study on Trajectory-Guided Extrapolation}
\label{Ablation Study on Trajectory-Guided Extrapolation}

The ablation study presented in Table \ref{tab:ablation_study1} provides valuable insights into the contributions of the key components within the Trajectory-Guided Extrapolation (TGE) framework, which are Success-Based Task Augmentation (SBTA) and Failure-Driven Task Exploration (FDTE).

Removing the entire TGE mechanism significantly impairs system performance, indicating that TGE plays a central role in driving effective extrapolation from trajectory data. Moreover, isolating the removal of either SBTA or FDTE reveals that both components contribute distinct and complementary benefits.

Among the two components, the analysis suggests that SBTA contributes more critically to performance improvements than FDTE. When SBTA is removed, performance degradation is more pronounced compared to the removal of FDTE. This indicates that leveraging successful trajectories for task augmentation has a greater overall impact on extrapolation quality than exploration driven by failure signals.

\begin{table}[t]
\centering
\caption{Comparison of NGPT metric across different methods on two benchmarks with the OS-Atlas backbone.}
\label{NGPT}
\begin{tabular}{lcccc}
\toprule
    \textbf{Method} & \textbf{Traj} & \textbf{$\Delta$Perf} & \textbf{$\Delta$Traj} & \textbf{NGPT} \\
\midrule
    \emph{VisualWebArena} &  &  &  &  \\
    \quad Baseline & 106 & 0 & 0 & / \\
    \quad BAGEL & 169 & -0.77 & 63 & -0.0122 \\
    \quad Self-Improve & 361 & -0.55 & 255 & -0.0022 \\
    \quad \textbf{CORE} & 413 & +2.53 & 307 & \textbf{0.0082} \\
\midrule
    \emph{AndroidWorld} &  &  &  &  \\
    \quad Baseline & 15 & 0 & 0 & / \\
    \quad BAGEL & 53 & +1.11 & 38 & 0.0292 \\
    \quad Self-Improve & 64 & +5.17 & 49 & 0.1055 \\
    \quad \textbf{CORE} & 60 & +7.75 & 45 & \textbf{0.1722} \\
\bottomrule
\end{tabular}
\end{table}

\subsection{Exploration Cost Efficiency Analysis}
\label{Exploration Cost Efficiency Analysis}

We define \textbf{NGPT} (Normalized Gain per Trajectory) for evaluating performance gains over exploration costs:
$$
\text{NGPT} = \frac{\Delta \text{Perf}}{\Delta \text{Traj}}
$$
where $\Delta \text{Perf}$ denotes the performance improvement over the baseline, and $\Delta \text{Traj}$ represents the additional trajectories used beyond the baseline. This metric measures the \textbf{efficiency} of each method in transforming extra exploration into performance gains. Table~\ref{NGPT} reports NGPT for OS-Atlas on both benchmarks. Our method achieves the highest NGPT score, indicating strong performance improvements at a relatively moderate exploration cost. This demonstrates that our framework is not only effective but also \textbf{cost-efficient}.



\subsection{CORE vs. Environment-Only Learning}
\label{CORE vs. Environment-Only Learning}

To examine whether our framework's performance gains stem merely from reliance on environment feedback, we conduct a controlled comparison between our framework and a baseline trained solely with \emph{environment feedback} as the reward signal. This baseline, denoted as \textbf{Env}, represents a purely method without structured guidance from the Strategy Graph or Label Functions.

\begin{figure}[t]
\centering
\includegraphics[width=0.45\textwidth]{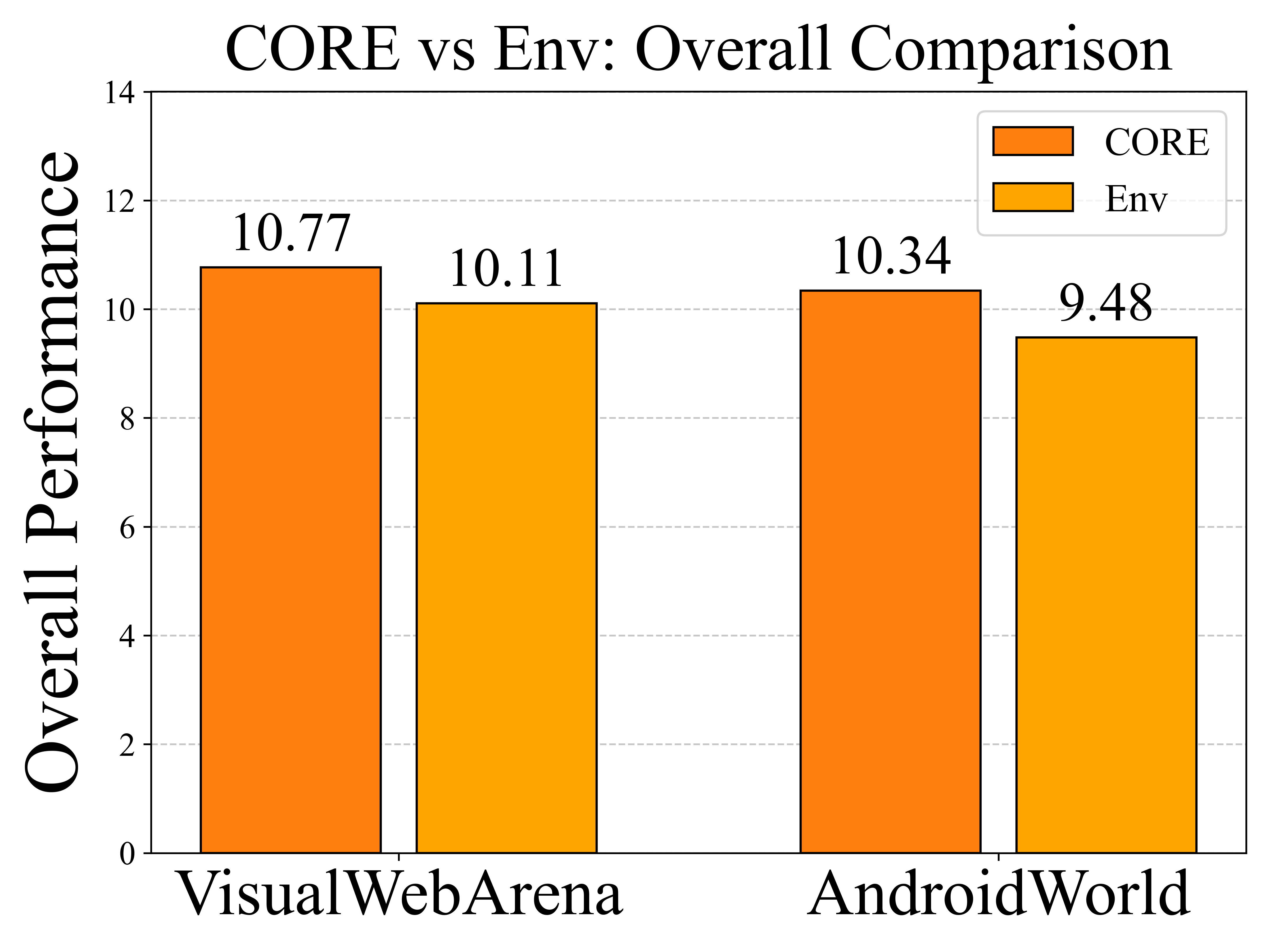}
\caption{Comparison between CORE and the environment-only baseline (Env) on \emph{VisualWebArena} and \emph{AndroidWorld}. CORE achieves consistently higher overall scores, demonstrating more effective utilization of environment feedback.}
\label{fig:core_env_comparison}
\end{figure}

As shown in Figure~\ref{fig:core_env_comparison}, our method consistently outperforms the Env baseline across both \textit{VisualWebArena} and \textit{AndroidWorld}. Specifically, CORE achieves higher overall scores of \textbf{10.77} vs.\ 10.11 on \textit{VisualWebArena}, and \textbf{10.34} vs.\ 9.48 on \textit{AndroidWorld}. These improvements highlight that while environment feedback remains an essential supervisory signal, CORE leverages it more effectively through structured reasoning and graph-based extrapolation, leading to more stable and sample-efficient learning.

\section{Prompt-guided Step Evaluations}
\label{Prompt-guided Step Evaluations}

\subsection{Evaluation of Key Step Identification}
\label{Evaluation of Key Step Identification}
The construction of the Strategy Graph relies heavily on the effective identification of key steps within the Semantic Code Abstraction process, which is critical for generating high-quality trajectories that align with environmental feedback. To evaluate the impact of this process, we conducted experiments across three iterations on two benchmarks. The performance of two models, OS-Atlas and GUI-R1, was assessed using four metrics: Accuracy, Precision, Recall, and F1-score, as shown in Table~\ref{tab:key_steps_table}. The results demonstrate consistently high performance, with all metrics averaging above 0.8 across both benchmarks. These results underscore the pivotal role of Key Step Identification in enabling the Strategy Graph to effectively filter and refine trajectories, ensuring robust and reliable performance in complex environments.

\begin{table}[t]
\centering
\caption{Performance metrics of OS-Atlas and GUI-R1 models across three iterations on \emph{VisualWebArena} and \emph{AndroidWorld} benchmarks, demonstrating the effectiveness of Key Step Identification in Strategy Graph construction. All metrics (Accuracy, Precision, Recall, and F1-score) consistently achieve high values, averaging above 0.8, highlighting the Effectiveness of the Key Step Identification process.}
\label{tab:key_steps_table}
\begin{tabular}{@{}clcccc@{}}
\toprule
\multirow{2}{*}{\textbf{Method}} & \multirow{2}{*}{\textbf{Metric}} & \multicolumn{3}{c}{\textbf{Iterations}} \\
\cmidrule{3-5}
& & \textbf{Iter. 1} & \textbf{Iter. 2} & \textbf{Iter. 3} \\
\midrule
    \multirow{4}{*}{\makecell{OS-Atlas \\ on \\ \textit{VisualWebArena}}} & Accuracy & 0.848 & 0.869 & 0.855 \\
     & Precision & 0.940 & 0.896 & 0.870 \\
     & Recall & 0.706 & 0.832 & 0.868 \\
     & F1-score & 0.806 & 0.863 & 0.869 \\
\cmidrule{1-5}
    \multirow{4}{*}{\makecell{GUI-R1 \\ on \\ \textit{VisualWebArena}}} & Accuracy & 0.849 & 0.863 & 0.852 \\
     & Precision & 0.943 & 0.889 & 0.895 \\
     & Recall & 0.800 & 0.807 & 0.847 \\
     & F1-score & 0.866 & 0.846 & 0.870 \\
\midrule
    \multirow{4}{*}{\makecell{OS-Atlas \\ on \\ \textit{AndroidWorld}}} & Accuracy & 0.925 & 0.900 & 0.891 \\
     & Precision & 0.846 & 0.800 & 0.849 \\
     & Recall & 0.733 & 0.857 & 0.918 \\
     & F1-score & 0.786 & 0.828 & 0.882 \\
\cmidrule{1-5}
     \multirow{4}{*}{\makecell{GUI-R1 \\ on \\ \textit{AndroidWorld}}} & Accuracy & 0.987 & 0.988 & 1.000 \\
     & Precision & 1.000 & 0.962 & 1.000 \\
     & Recall & 0.909 & 1.000 & 1.000 \\
     & F1-score & 0.952 & 0.980 & 1.000 \\
\bottomrule
\end{tabular}
\end{table}

\subsection{Evaluation of Label Function Synthesis}
\label{Evaluation of Label Function Synthesis}

The Label Function Synthesis module aims to automatically generate executable reward functions that verify whether an agent correctly completes a specific key step. We employ the \textbf{Qwen2.5-Coder-3B-Instruct} model to generate Label Functions for each key step, performing five synthesis attempts per instance. The synthesized functions are then executed and validated against expert trajectories to determine their functional correctness.  

We report three metrics: (1) \textbf{OSR} (Overall Success Rate): the proportion of successful syntheses within five attempts, (2) \textbf{FTSR} (First-Try Success Rate): the proportion of Label Functions that succeed on the first attempt, and (3) \textbf{ESP} (Early Success Position): the average position of the first successful synthesis. Results on the \textit{VisualWebArena} and \textit{AndroidWorld} benchmarks are summarized in Table~\ref{tab:label_function_table}.  

Across both benchmarks, all models achieve remarkably high OSR values (above 0.91 on \textit{VisualWebArena} and 1.00 on \textit{AndroidWorld}), indicating the strong reliability of the synthesis process. The high FTSR (ranging from 0.81 to 0.99) and low ESP (close to 1.0) further demonstrate that most Label Functions are correct on the first or second attempt, requiring minimal retries. These findings confirm that the proposed Label Function Synthesis method can robustly and efficiently infer executable step-level reward functions from semantic abstractions, ensuring accurate supervision for downstream trajectory evaluation.

\begin{table}[t]
\centering
\caption{Performance of Label Function Synthesis evaluated on \emph{VisualWebArena} and \emph{AndroidWorld} benchmarks. OSR: Overall Success Rate, FTSR: First-Try Success Rate, ESP: Early Success Position. All values indicate high reliability and efficiency of the synthesis process.}
\label{tab:label_function_table}
\begin{tabular}{@{}llccc@{}}
\toprule
\textbf{Benchmark} & \textbf{Model} & \textbf{OSR} & \textbf{FTSR} & \textbf{ESP} \\
\midrule
\multirow{2}{*}{\textit{VisualWebArena}} 
& OS-Atlas & 0.9150 & 0.8095 & 1.1710 \\
& GUI-R1   & 0.9179 & 0.8446 & 1.1246 \\
\midrule
\multirow{2}{*}{\textit{AndroidWorld}} 
& OS-Atlas & 1.0000 & 0.9899 & 1.0101 \\
& GUI-R1   & 1.0000 & 0.9674 & 1.0435 \\
\bottomrule
\end{tabular}
\end{table}

\subsection{Evaluation of New Intent Generation and Refinement}
\label{Evaluation of New Intent Generation and Refinement}

The New Intent Generation and Refinement module aims to exploit the potential of failed trajectories by inferring alternative task intents that better align with the agent's observed behaviors. To evaluate this process, we adopt an \textbf{LLM-as-Judge} protocol. We prompt \textbf{Qwen2.5-VL-32B-Instruct} with a trajectory, the original intent, and the newly generated intent, asking which intent better matches the given trajectory in terms of logic, relevance, and goal alignment. The model's response distribution is analyzed, and we report the percentage of cases where it prefers the new intent over the original one, as shown in Table~\ref{tab:new_intent_table}.

\begin{tcolorbox}[
    colback=gray!10!white,
    colframe=gray!50!black,
    title=Prompt for LLM-as-Judge,
    fonttitle=\bfseries,
    boxrule=0.5mm,
    arc=2mm,
    width=\columnwidth,
    breakable,
    before skip=2mm,
    after skip=2mm,
    left=3pt,
    right=3pt,
    top=3pt,
    bottom=3pt
]
{\ttfamily\small
You are given a trajectory and two possible intents. Decide \textbf{which intent better matches the trajectory}. Consider logic, relevance, and goal alignment.\\
Trajectory: <<trajectory>>\\
Intent1: <<original\_intent>>\\
Intent2: <<new\_intent>>\\
Output: Only respond with one of the following:\\
`Intent 1` / `Intent 2` / `Cannot decide`
}
\end{tcolorbox}

Results show that the refined intents are judged as better aligned with the trajectories in \textbf{78.92\%} of cases on \textit{VisualWebArena} and \textbf{82.81\%} on \textit{AndroidWorld}. This consistent preference for the refined intents indicates that our generation and refinement pipeline effectively extracts meaningful alternative goals from failed experiences, thereby transforming otherwise unusable data into high-quality supervision signals that improve generalization and robustness.

\begin{table}[t]
\centering
\caption{LLM-as-Judge Evaluation of New Intent Generation and Refinement. The ratio denotes the percentage of cases where the new intent is preferred over the original intent. Higher ratios indicate better alignment between new intents and trajectories.}
\label{tab:new_intent_table}
\begin{tabular}{@{}llc@{}}
\toprule
\textbf{Benchmark} & \textbf{Model} & \textbf{Ratio (\%)} \\
\midrule
\multirow{3}{*}{\textit{VisualWebArena}} 
& OS-Atlas & 78.35 \\
& GUI-R1   & 79.47 \\
& \textbf{Average} & \textbf{78.92} \\
\midrule
\multirow{3}{*}{\textit{AndroidWorld}} 
& OS-Atlas & 90.16 \\
& GUI-R1   & 76.12 \\
& \textbf{Average} & \textbf{82.81} \\
\bottomrule
\end{tabular}
\end{table}

\section{Case Studies}
\label{Case Studies}
\subsection{Case Study of Label Functions}
\label{Case Study of Label Functions}

We present illustrative case studies demonstrating the design of Label Function used to verify the execution of a specific key step within one task. Each case provides a key step description and showcases how the corresponding label function programmatically validates the success of the key step based on trajectory data. These examples reflect the versatility and robustness of our Label Function design across different settings.

\subsubsection{Case 1 on \textit{VisualWebArena}}
\begin{tcolorbox}[
    colback=gray!10!white,
    colframe=gray!50!black,
    title=Case 1 on \textit{VisualWebArena},
    fonttitle=\bfseries,
    boxrule=0.5mm,
    arc=2mm,
    width=\columnwidth,
    breakable,
    before skip=2mm,
    after skip=2mm,
    left=3pt,
    right=3pt,
    top=3pt,
    bottom=3pt
]
{\ttfamily\small
    \textbf{Key Step Description:} Click the link 'Add to Wish List' for the 'PEACE NEST Lightweight Down and Feather Fiber Throw Blanket Soft Couch Throw for Indoor and Outdoor Use, 50x70, Navy Blue'.\\

    \textbf{Observation:}\\
    \begin{center}
    \vspace{-10pt}
    \includegraphics[width=0.6\linewidth]{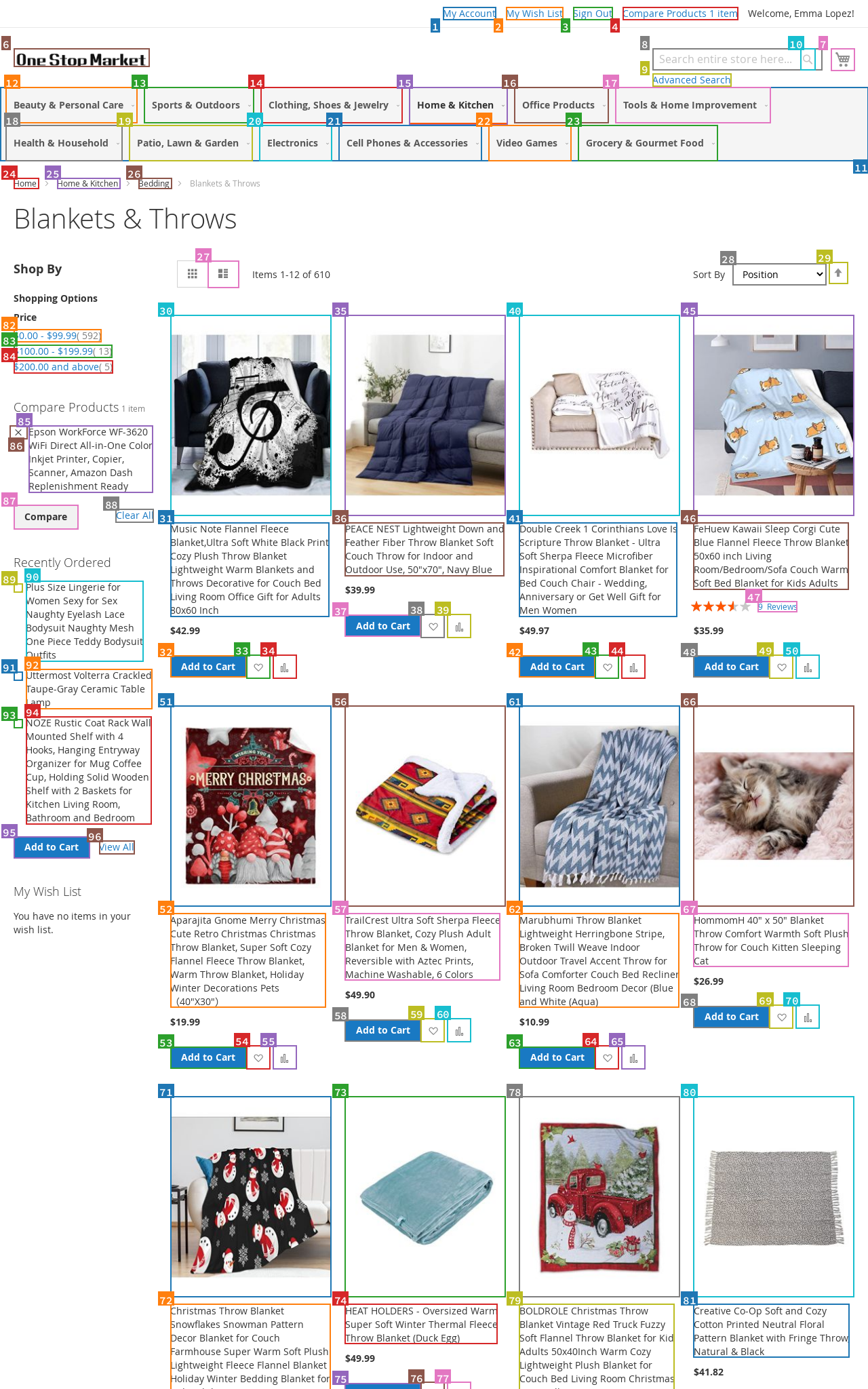}
    \vspace{5pt}
    \end{center}
    
    \textbf{Label Function:}

    \begin{lstlisting}
from Function_APIs import *
def verify_function(trajectory, stop_page_url):
    # Check if the 'Add to Wish List' link was clicked
    if not validate_click_or_hover_action(trajectory, 'click', 'A', 'Add to Wish List'):
        return False
    # Check if the item 'PEACE NEST Lightweight Down and Feather Fiber Throw Blanket Soft Couch Throw for Indoor and Outdoor Use, 50x70, Navy Blue' was added to the wishlist
    if not validate_item_in_wishlist(trajectory, 'PEACE NEST Lightweight Down and Feather Fiber Throw Blanket Soft Couch Throw for Indoor and Outdoor Use, 50x70, Navy Blue'):
        return False
    # Return True if all conditions were satisfied
    return True
# Execute and return result
result = verify_function(trajectory, stop_page_url)
    \end{lstlisting}
}
\end{tcolorbox}
\vspace{10pt}

In this case, the task goal is ``Add the navy blue one in the second column to my wish list", and the key step involves clicking the ``Add to Wish List" link for the product `PEACE NEST Lightweight Down and Feather Fiber Throw Blanket Soft Couch Throw for Indoor and Outdoor Use, 50x70, Navy Blue'. The Label Function first verifies whether the correct click action was performed, and then checks if the specified item was successfully added to the wishlist. This two-stage verification ensures both the correct action and its intended effect are validated, enhancing robustness.

\subsubsection{Case 2 on \textit{VisualWebArena}}
\begin{tcolorbox}[
    colback=gray!10!white,
    colframe=gray!50!black,
    title=Case 2 on \textit{VisualWebArena},
    fonttitle=\bfseries,
    boxrule=0.5mm,
    arc=2mm,
    width=\columnwidth,
    breakable,
    before skip=2mm,
    after skip=2mm,
    left=3pt,
    right=3pt,
    top=3pt,
    bottom=3pt
]
{\ttfamily\small
    \textbf{Key Step Description:} Stop the task with answer: '4200 calories'.\\

    \textbf{Observation:}\\
    \begin{center}
    \vspace{-10pt}
    \includegraphics[width=0.7\linewidth]{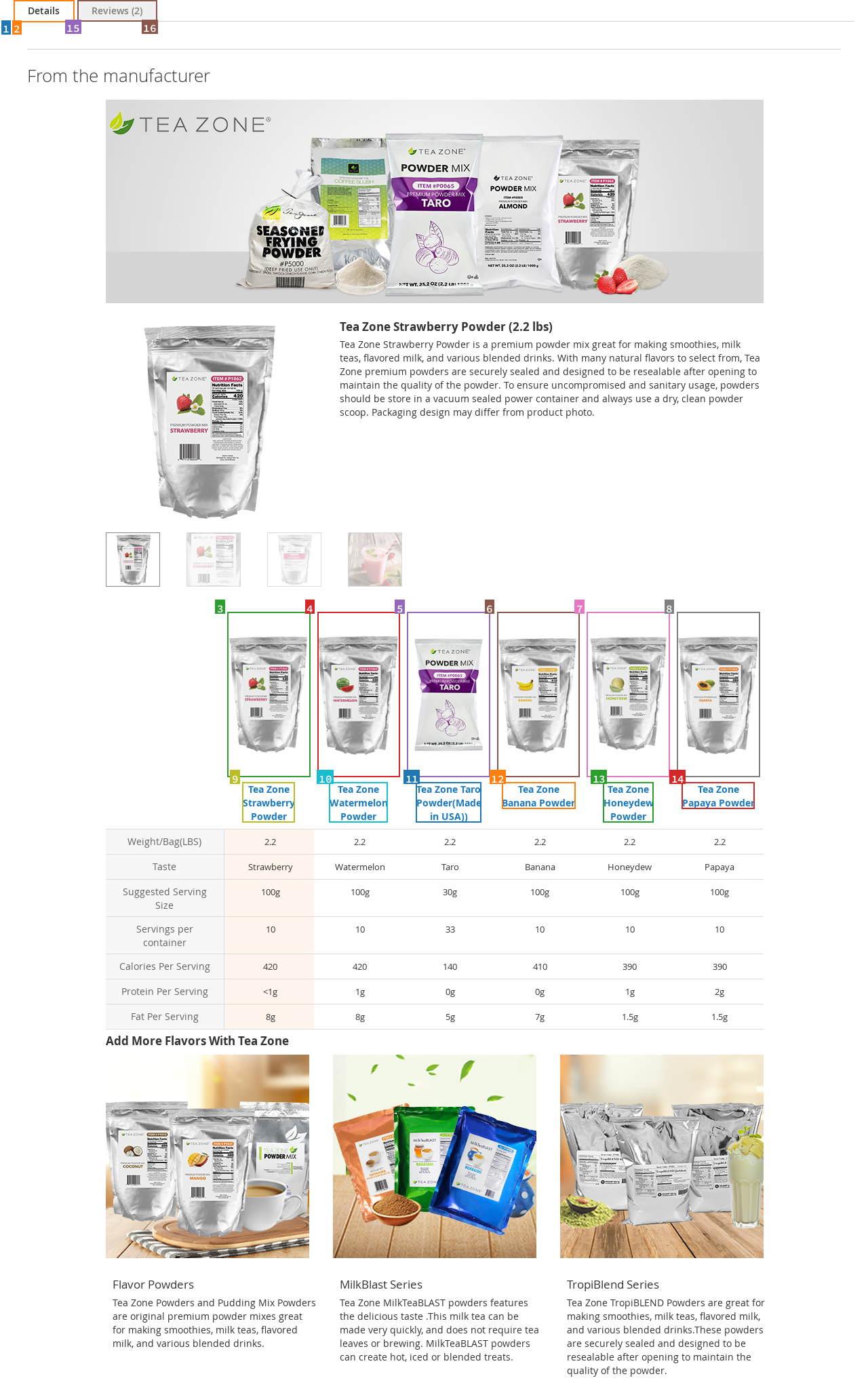}
    \vspace{5pt}
    \end{center}
    
    \textbf{Label Function:}

    \begin{lstlisting}
from Function_APIs import *
def verify_function(trajectory, stop_page_url):
    if not validate_stop_action(trajectory, '4200 calories'):
        return False
    return True
# Execute and return result
result = verify_function(trajectory, stop_page_url)
    \end{lstlisting}
}
\end{tcolorbox}

In this case, the task goal is ``How many calories are in this item per container?", and the key step is to stop the task with answer: `4200 calories'. The label function checks for a stop action accompanied by the precise answer ``4200 calories". It verifies the task's success by directly matching the final textual outcome, which is appropriate for information retrieval tasks involving precise values.

\subsubsection{Case 1 on \textit{AndroidWorld}}
\begin{tcolorbox}[
    colback=gray!10!white,
    colframe=gray!50!black,
    title=Case 1 on \textit{AndroidWorld},
    fonttitle=\bfseries,
    boxrule=0.5mm,
    arc=2mm,
    width=\columnwidth,
    breakable,
    before skip=2mm,
    after skip=2mm,
    left=3pt,
    right=3pt,
    top=3pt,
    bottom=3pt
]
{\ttfamily\small
    \textbf{Key Step Description:} Type text 'Clock' into the target text field 'Search apps, web and more'.\\

    \textbf{Observation:}\\
    \begin{center}
    \vspace{-10pt}
    \includegraphics[width=0.6\linewidth]{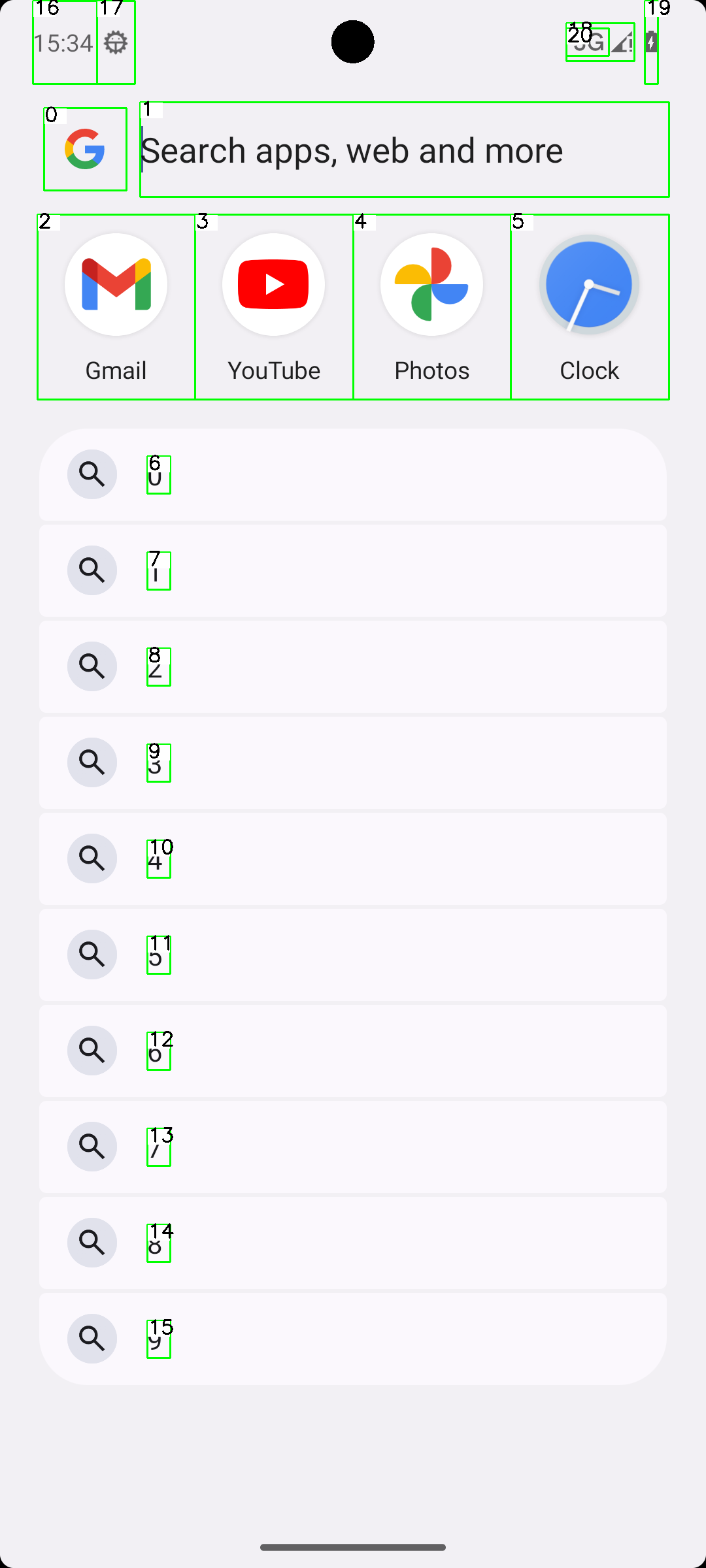}
    \vspace{5pt}
    \end{center}
    
    \textbf{Label Function:}

    \begin{lstlisting}
from Function_APIs import *
def verify_function(trajectory):
    if not validate_type_action(trajectory, 'Clock', target_text_field='Search apps, web and more'):
        return False
    return True
# Execute and return result
result = verify_function(trajectory)
    \end{lstlisting}
}
\end{tcolorbox}

In this case, the task goal is ``Run the stopwatch.", and the key step is to type text `Clock' into the target text field `Search apps, web and more'. The Label Function confirms the text input ``Clock" into a designated search field. This verifies not just the content typed but its correct placement within the intended UI element, which is essential for validating navigational steps involving search-based interaction.

\subsubsection{Case 2 on \textit{AndroidWorld}}
\begin{tcolorbox}[
    colback=gray!10!white,
    colframe=gray!50!black,
    title=Case 2 on \textit{AndroidWorld},
    fonttitle=\bfseries,
    boxrule=0.5mm,
    arc=2mm,
    width=\columnwidth,
    breakable,
    before skip=2mm,
    after skip=2mm,
    left=3pt,
    right=3pt,
    top=3pt,
    bottom=3pt
]
{\ttfamily\small
    \textbf{Key Step Description:} Click on a UI element 'Pro Expense' on the screen.\\

    \textbf{Observation:}\\
    \begin{center}
    \vspace{-10pt}
    \includegraphics[width=0.5\linewidth]{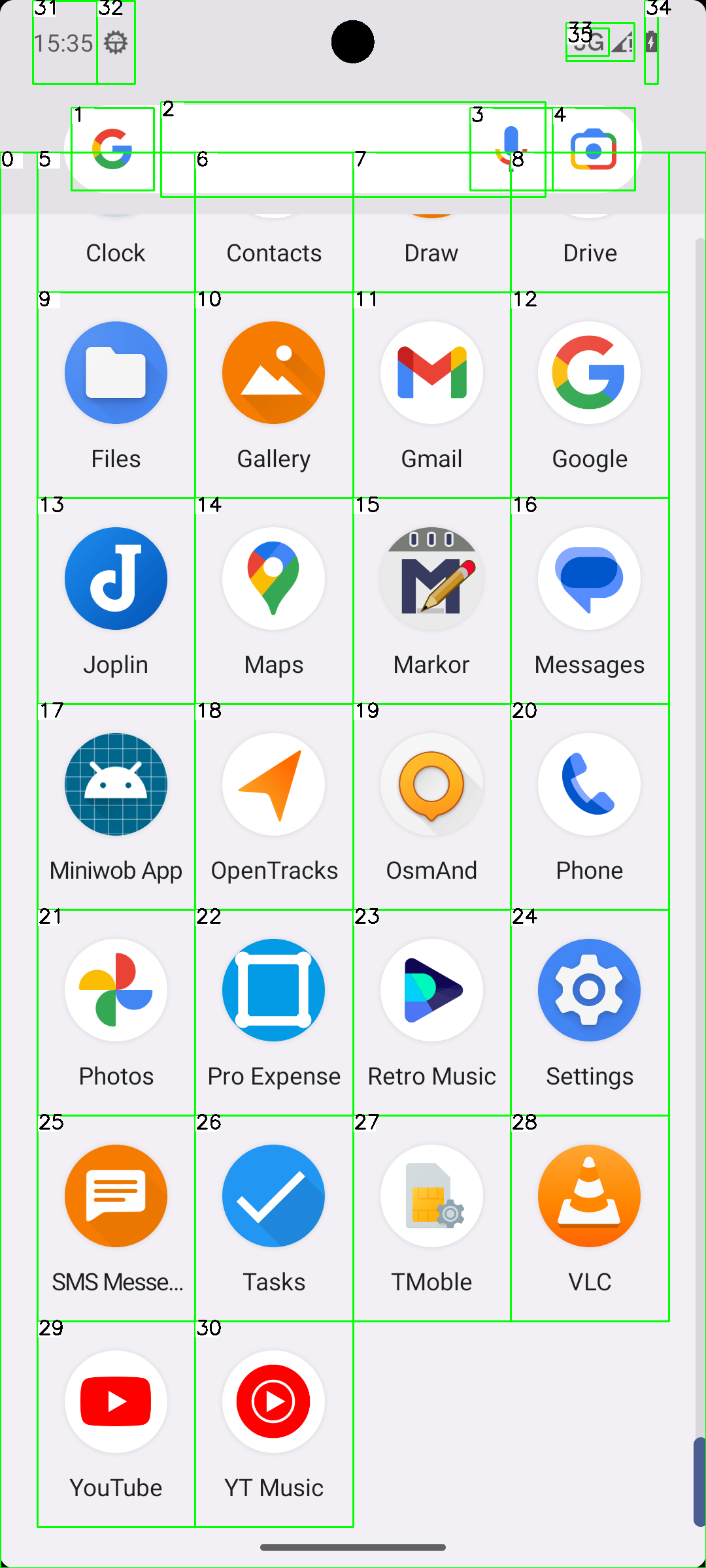}
    \vspace{5pt}
    \end{center}
    
    \textbf{Label Function:}
    \begin{lstlisting}
from Function_APIs import *
def verify_function(trajectory):
    if not validate_click_action(trajectory, 'Pro Expense'):
        return False
    return True
# Execute and return result
result = verify_function(trajectory)
    \end{lstlisting}
}
\end{tcolorbox}

In this case, the task goal is ``Delete the following expenses from pro expense: Rental Income.", and the key step is to click on a UI element `Pro Expense' on the screen. The Label Function validates whether the user clicked on the ``Pro Expense" UI element. Since this step is often preparatory for more complex interactions (like deletion), ensuring the correct interface element was targeted is key for downstream task success.

\end{document}